# HIERARCHICAL MODELING

# OF MULTIDIMENSIONAL DATA

# IN REGULARLY DECOMPOSED SPACES

## Tome 1 : MAIN PRINCIPLES

## (1984 – 1988)

- 2016 -

**Olivier Guye**

# Table of Contents

# Table of Figures

# Table list

# Introduction

The works presented hereinafter have been carried out in the framework of a mid-term study initiated by the Centre Electronique de l'Armement and led by ADERSA, a company of research under contract (authorization ANVAR n°B7911050W).

ADERSA was at those times a small and middle size enterprise carrying out applied research works in the field of continuous process control. It has been created by Dr Jacques Richalet, who is considered as one of the pioneers of model-based predictive control and who received the Nordic Process Control Award in 2007 for his significant contributions in process control delivered during all his scientific career. ADERSA has so developed several different methodologies in the field of model-based predictive control that can apply to fast process control (closed-loop control in robotics) as well as to slow processes (batch processing in the petro-chemical industry). In addition to these two basic fields, ADERSA has developed some other some other ones linked to research sectors as economic modeling, failure diagnosis, image analysis and problem solving, relying on similar techniques.

On its own side the CELAR was already implied in the design of flight simulators and foresaw that next command and control systems should rely on a more advanced modeling of operation theaters and would need the use of new tools for problem solving in order to carry out missions of different natures:

- the knowledge of modeling multidimensional numerical objects determines the design of modern systems in a lot of domains ;
- design, computer-aided manufacturing robotics, image analysis and synthesis, pattern recognition, decision making, cartography, databases ;
- the needs in capture, processing, visualization and transmission of information of bi- or tridimensional nature are known, but also exist for multidimensional data.

On its side, ADERSA has just developed a new modeling technique relying on a piecewise multiple regression based on a recursive division of a data set orthogonally to the main inertia axis with the hyperplane passing by the gravity center of the points cloud. The result of this dividing process organized the data to be modeled into a binary tree where the neighboring data are gathered into sub-sets modeled by a single linear model respecting a given approximation error.

In the framework of the proposed study, the CELAR was willing that the interest was rather focused on regular division techniques that are easier to developed and that may have a wider usage spectrum than the piecewise multiple regression :

- the principles of regular hierarchical decomposition was already applied with success in bi- and tridimensional spaces in the form of quadtrees and octtrees ;
- it seemed that these methods can be extended to spaces of any dimension for indexing data in a multidimensional database by using a kind of kd-tree.

These study works have been carried out in the course of two research contracts CELAR-ADERSA n° 005/41/84 and n°004/41/88 and then have been at the origin of several other works dealing with the evaluation of this methodology in different application domains.

The results shown in the present document are mainly dealing on the works performed during the first study. They have been the subject of a publication in two parts.

# I - Background

The usage of the "divide and conquer" paradigm has allowed to develop algorithms satisfying optimal bounds for some classical problems as sorting or the computation of a convex hull ([KNUTH 73], [AHO 74], [PREPARATA 77], [PREPARATA 84]). It consists in decomposing a problem that can be directly solved into sub-problems and iterating this approach until that all the problems have been solved. When the problem to be solved is divided in two others, the data structure used for data management is a binary tree.

For solving the problem of the hidden parts removal when a tridimensional object is displayed on a flat screen, WARNOCK has built a quaternary tree in applying this approach for managing the visible parts of an object ([WARNOCK 69], [SUTHERLAND 74B], [NEWMAN 75]). In this respect, he is considered as the inventor of this data structure; he is more known for his participation to the design of the typographic language POSTSCRIPT and to the foundation of ADOBE enterprise.

The first main reference which is usually highlighted is the paper of KLINGER et DYER ([KLINGER 76]) dealing with the analysis of this data structure and its properties of symmetry. These works about the identification of symmetries will go in depth with the collaboration of ALEXANDRIDIS ([ALEXANDRIDIS 78, 84]), and will initiate significant works about the search of adjacencies and the labeling of connected components

The remarkable properties of quaternary trees are then highlighted, it is a data structure :

- whose compression rate is at least equal to run length coding ([DYER 82]);
- that can be easily used for implementing Boolean operators, a scaling by step of 2, a translation and a rotation by step of 90° ([OLIVER 83a, 86b]).

The quaternary trees are implemented according two different representation schemes on computers: linked lists or linear codes. It is in this last scheme that the smallest data size can be reached, but also where the algorithmic constraints are the most severe (the access to one piece of information needs to visit the whole data code).

The linear codes can be divided into two classes:

- tree codes, where all the nodes of the tree are coded according to a given path;
- leaf codes where only terminal nodes are coded and gathered into a single collection.

According their production, two researchers can be distinguished: SAMET of Maryland University and GARGANTINI in Canada. GARGANTINI has studied quaternary trees and their tridimensional extension, the octernary trees, modeled by leaf codes ([GARGANTINI 82a - 86b]). SAMET has mainly focused his efforts on quaternary trees modeled by linked lists ([SAMET 79 - 85e]).

These two researchers have mainly solved the problem of the adjacency search and the labeling of connected components. SAMET has developed a procedure for computing the median axes of a set ([SAMET 82b], [SAMET 83]) and has implemented a cartographic information system based on these last data structures ([SAMET 84c]).

Under the supervision of ROSENFELD, it has been established at Maryland University an authentic school on the study of hierarchical data structures in image analysis ([ROSENFELD 80 -84b]). Many searchers have so collaborated with SAMET: DYER in adjacency search ([DYER 82]), RANADE in filtering and attributes calculation ([RANABE 81a - 82]), SHNEIER ([SCNEIER 81a - 8 lb]) and TAMMINEN ([TAMMINEN 84a - 84b]).

It is not very convenient to apply linear transforms on trees: scaling, translation and rotation of any angle. HUNTER et STEIGLITZ have analyzed the issue ([HUNTER 79a - 79b]) and MEAGHER will solve it by including also the perspective for implementing a display system for tomographic images ([MEAGHER 80 - 82c]). These works will be carried on leaf codes by VAN LIEROP ([VAN LIEROP 86]). They allow to provide a new tool for modeling the configuration space of a mobile robot and to solve the problem of obstacle avoidance ([FAVERJON 84], [LOZANO-PEREZ 85], [HONG 85]).

YAU and SRIHARI have studied the reconstruction of tomographic images from parallel slices, the extraction of randomly placed slices, the building of a convex hull ([SRIHARI 81], [YAU 81 - 84]).

BENTLEY successfully dealt with the tree-like management of multidimensional numerical databases ([BENTLEY 75 - 80]). YAU and SRIHARI analyzed the possibility to model images of any dimension ([YAU 83]). CHAUDHURI studied the multidimensional trees as a classification and pattern recognition technique ([CHAUDHURI 85]). In statistical data analysis, several comparisons have been made between the PEANO-HILBERT scanning and partitioning methods.

In image compression, higher compression rates have been reached in comparison with previous methods ([PAVEL 85], [KUNT 87]). In finite elements, quaternary and octernary trees have been used in order to automatically generate meshes ([YERRI 83], [SHEPARD 85a - 85b]). Several authors pointed out that this modeling structure enables to parallelize algorithms based on them.

Two synthesis books describe the results obtained with quaternary or octernary trees and pyramids ([TANIMOTO 80], [ROSENFELD 84b]).

# II – Hierarchical modeling of numerical data

## *II.1 - Presentation*

The principle consists in representing the numerical object to be modeled by a fixed-shape box at the highest level, then in dividing this box into sub-boxes according to a procedure defined in advance and in applying recursively this dividing principle upon each sub-box either until finding boxes of similar value, or until having reached the desired modeling precision.

Let us consider a planar binary image, it can be divided into four quadrants:

- north-west,
- north-east,
- south-east,
- south-west.

Each quadrant may have a binary uniform color (0 or 1) or not. The quadrants that have not got a uniform color are once more divided.

The underneath figure illustrates this dividing process and is showing that a graph of outer degree 4 (children number of a node) is generated in such a way: a quaternary tree.

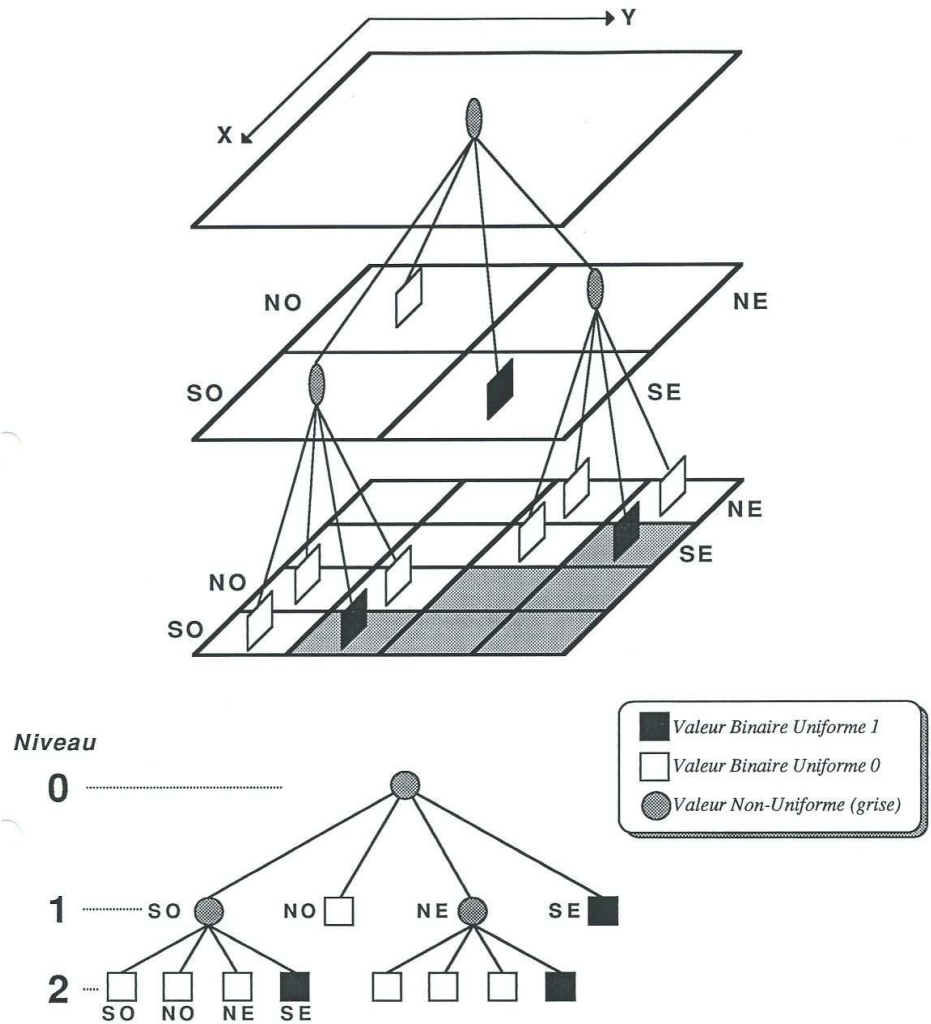

**Figure 1 : Quaternary tree of a planer binary image**

The same building process can be adapted to tridimensional binary images so as to get an octernary tree.

The quadrants become octants so as to take in account a complementary division made along the object height, they are then eight of them:

- bottom-north-west,
- top-north-west,
- •
- •
- •
- bottom-south-east.

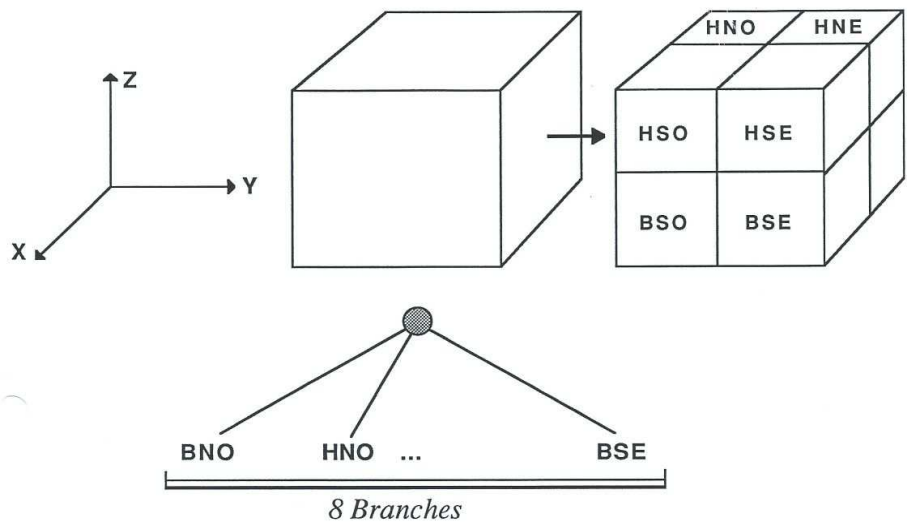

**Figure 2 : Decomposition of a tridimensional binary image into octants**

It can be noticed that this decomposition can be generalized to a number $k \geq 1$ of dimensions. It will be then produced a tree of outer degree $2^k$. For instance, a mono-dimensional binary signal can be modeled by a binary tree, a temporal series of tridimensional binary images by a tree of degree $16$.

The objects modeled by a $2^k$-tree should be representable in a universe where each of the k dimensions are discretized over the same sub-set of $N : \{0,...,2^r -1\}$, where $r$ is the maximum level of filiation reachable in the tree. The $2^k$-trees are complete trees.

## II.2 – Sequential allocation: the linear lists

Two approaches are the most often used.

The first one, a tree code, is providing the best information compression ratio. Coding is performed by sequentially registering the color of each node of the tree according to a predefined traversal. For instance if the colors white, black and grey are coded 0, 1, 2 (using two coding bits), the 4-tree of figure 1, will be expressed as the characters' strings:

- 0 0 0 1 2 0 0 0 0 1 2 1 2    in post-order,
- 2 2 0 0 0 1 0 2 0 0 0 1 1    in depth-first

The main disadvantage of this data structure is that it is necessary to read the coded string from the beginning in order to reach any node.

The second approach, used in sequential allocation, consists in leaf codes. Here the coding does not ever favor the node color, but its position in the discretization space. The resulting coding provides a less good compression than tree codes, but enables faster access by performing a search based on a dichotomous sort above the so generated list and allowed by the order relation induced by the coding.

The coding alphabet used for the 4-trees is (0, 1, 2, 3, X}, where 0, 1, 2, 3 refer to the four quadrants of decomposition and X the un-development of a terminal node at an intermediate level inside the tree. Code generation is performed by duplication of the code of a non-terminal node to be decomposed and concatenation of the quadrant number of decomposition. By noticing that the code of terminal nodes are including those of their fathers, these ones are not kept in the resulting list. Restarting from the example of figure 1, the leaf code produced according to this method would be, if (SO, NO, NE, SE} is coded by (0, 1, 2, 3) :

- (03, 23, 3X}  by coding black nodes,
- (00, 01, 02, 1X, 20, 21, 22}  by coding white nodes.

Each one of these lists are ordered according to used quaternary alphabet and the applied function of composition. The processed trees are no more complete At the opposite, each node is including in its coding the path that links it to the tree root. The quaternary coding has an octernary equivalent for octernary trees

## II.3 – Indexed allocation: the linked lists

In a tree, there are two partial order relations:

- filiation order,
- and sibling order.

The filiation order is usually implemented at first in tree-like lists.

If it is planned to manage a tree according this single order, it is necessary to allocate to each tree node, as many filiation links as the maximum number of children that it may have.

For enabling only downwards moves, it will be needed:

- two successors for a binary tree, usually named left and right sons
- four successors for a 4-tree,
- $2^k$ successors for a $2^k$-tree.

For enabling also upwards moves, only one complementary pointer is needed whatever can be the tree: a father link.

Restarting from the example of figure 1, it would be got the linked lists shown in figure 3. On this example, it can be noticed that more than a half of memory usage is dedicated to the links of the children of the terminal nodes. These data do not hold any meaning power and consist in a loss of memory space but is necessary for insuring the consistency of the building. Usually, terminal nodes are implemented in another way in order to minimize memory occupation.

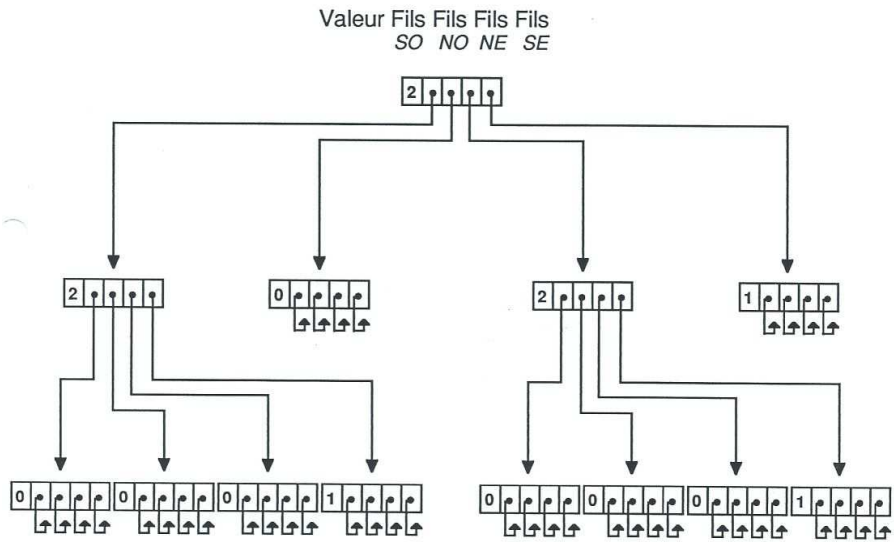

**Figure 3 : Example of linking based on simple filiation**

## II.4 - Emulation of a $2^k$-tree by a binary tree

The filiation order relation enables to represent a $2^k$-tree by a binary tree. It is no more consisting in performing a decomposition quadrant by quadrant or octant by octant, but a median division according each dimension of the space, each of them being processed sequentially one after another one..

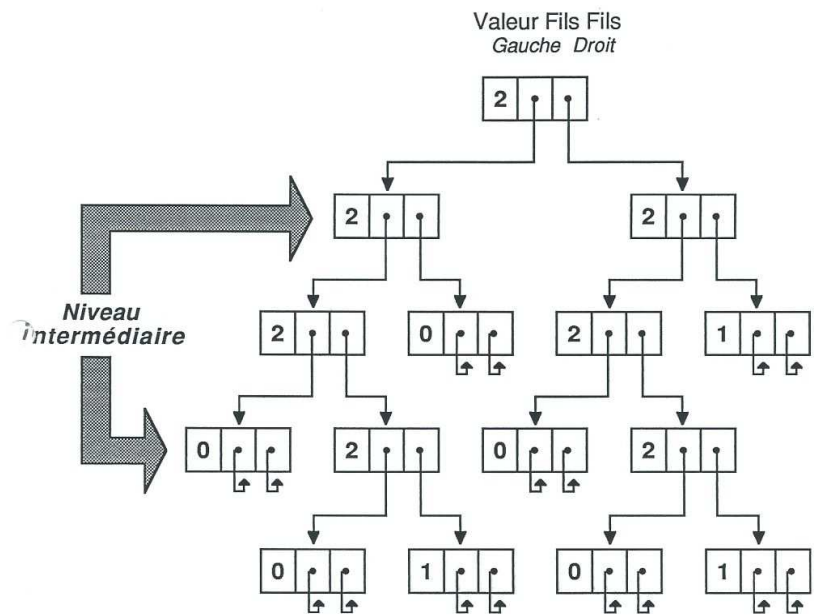

**Figure 4 : Representation of a quaternary tree by a binary tree**

## II.5 – Memory requirements and arithmetic features of 2^k-trees

A $2^k$-tree models a universe of dimension k. If $r$ is the filiation maximum level in the tree, each of its dimensions are then mapped over the sub-set of N : $\{0,...,2^r -1\}$.

An often used numbering for indexing nodes in a tree preserving the filiation and sibling orders is the following one:

node index at the level of depth $r$ in the tree :
$$\sum_{i=1}^{r} \left(2^{k(i-1)} s(i)\right),$$
where $s(i) \in \{0,...,2^k -1\}$    sibling order of a node among its brothers,
where $\{s(i), i = 1,...,r\}$    is the tree traversal enabling to reach the target node at level $r$,
and where $2^{k(i-1)}$    is the filiation order at level $i$.

This numbering creates a linear order over the tree nodes. It is induced by a breadth-first traversal on a complete tree and linearly counted down to level $r$.

It can be noticed that the sum can be interpreted as a concatenation and that the deducted arithmetic is an arithmetic relying on variable-length numbers. So, the first child 0 of the root and the traversal 00...0 down to level $r$ do not represent the same value. The tree code 0 represents the empty set

The tree code 22210 or the traversal 0000 represents the value 0 or the interval of values $[0, 2^{-4}[$ at an infinite precision. GARGANTINI shows that this numbering enables to directly find the coordinates of the vertex of the hypercube which is the closest from the origin of the universe and consequently its center.

A hypercube located in a $2^k$-tree at level $r$ gets $\dfrac{1}{2^{kr}}$ as volume value, when the modeled universe is corresponding to the normalized hypercube $[0,1]^k$.

The maximum number of nodes included in a $2^k$-tree developed down to level $r$ is:
$$\sum_{i=0}^{r} 2^{ki} = \frac{2^{k(r+1)} - 1}{2^k - 1} \approx 2^{kr}$$

When this $2^k$-tree is represented by a binary tree, the maximum number of nodes becomes:
$$\sum_{i=0}^{kr} 2^i = 2^{kr+1} - 1$$

It can be noticed that the binary tree will have nearly twice nodes than a $2^k$-tree. Knowing that in a binary tree, the children number is always two, a quaternary tree requires nearly the same memory size than its representation by a binary tree and, for any space of dimension k > 3, a binary tree will need less memory than a $2^k$-tree implemented using a linked list.

## II.6 – Geometric interpretation of $2^k$-trees

The $2^k$-trees are applied on multidimensional binary spaces. SRIHARI is seeing them as the representation of the indicator function of objects belonging to the space to be modeled.

After having decomposed the corresponding space, this one appears as a set of unitary elements V above which the modeled object is represented by the function $f:\{v\} \rightarrow \{0,1\}$ such as:

- the object is the set $S = \{v / f(v) = 1\}$,
- the background of the universe on which the object is described is the set $\overline{S} = \{v / f(v) = 0\}$.

So as to take in account multi-valued data, the representation can be extended to the following formalism:

let $f_1, f_2, ..., f_n$ functionals defined on $(u_1, u_2, ..., u_m)$, it can be defined the indicator function:
$$\delta : (u_1, u_2, ..., u_m, f_1(u_1, u_2, ..., u_m), ..., f_m(u_1, u_2, ..., u_m)) \rightarrow \{0,1\}$$
where the value 1 is taken when the (m + n)-uplet exists and the value 0 at the opposite.

The set $S = \{v / \delta(v) = 1\}$ will be the modeled numerical object and $\overline{S} = \{v / \delta(v) = 0\}$ the universe background.

The tree-like modeling assumes that it is possible to regularly divide the object according each of these dimensions:

- either on the sub-set of the natural integers $\{0, 1, ..., 2^r - 1\}$,
- either on the sub-set of rational numbers $\left\{0, \frac{1}{2^r}, ..., \frac{2^r - 1}{2^r}\right\}$

or any variation between these two representations.

The function $\delta$ is then defined on the set approximated at the precision $r$ $\left\{0, \frac{1}{2^r}, ..., \frac{2^r - 1}{2^r}\right\}^{n+m}$ of the unitary hypercube $[0,1[^{n+m}$.

By distinguishing no more the functionals from the variables on which they are applying, a $2^k$-tree will describe an object belonging to the digital universe $\left\{0, \frac{1}{2^r}, ..., \frac{2^r - 1}{2^r}\right\}^k$

So a multilevel bi-dimensional image will have as indicator function:

$$\delta(x, y, lu\min escence) \rightarrow \{0,1\} \text{ and a 8-tree as model.}$$

Likewise a colored bi-dimensional image will be similarly modeled by a $2^5$-tree so as to take in account its three fundamental components.

To emulate a $2^k$-tree with a binary tree results to some modifications on this scheme. Actually the support of the indicator function developed at the precision $r$ is still not represented by the set:

$$\left\{0, \frac{1}{2^r}, ..., \frac{2^r-1}{2^r}\right\}^k, \text{ but by } \left\{0, \frac{1}{2^{kr}}, ..., \frac{2^{kr}-1}{2^{kr}}\right\}$$

on the unitary segment $[0,1[$.

It is due to the fact that the division is not applied in parallel on the k dimensions of the space, but sequentially one dimension after the other one. So the crossing from the precision $r$ to $r+1$ is performed by examining the $2^k$ possible values of:

$(u_1/2, u_2/2, ..., u_k/2)$ after having gone through those of the previous levels.

On a binary tree, this same operation consists in evaluating the number:

$$u = u_1/2 + u_2/2^2 + ... + u_k/2^k.$$

A node index located at the precision $r$ in the tree will get as value in a $2^k$-tree:

$$\sum_{i=1}^{r}\left(2^{k(i-1)} s(i)\right)$$

where $s(i) \in \{0, ..., 2^k - 1\}$ is the sibling order of a node among its brothers;
and where $2^{k(i-1)}$ is the filiation order at level $i$.

In the case of a binary tree, the node index at the depth $r$ will have as value:

$$\sum_{i=1}^{r}\left(2^i s(i)\right) \text{ where } s(i) \in \{0,1\}$$

## II.7 – Editing distance between trees

The problem is settled in a lexicographic manner by calculating the minimal cost enabling to transform a tree A into a new one B. At each tree node is associated a label, and for editing a tree, only three operations are allowed:

- change the label of a tree node,
- insert a sub-tree into a tree,
- delete a sub-tree from a tree.

Each operation gets a positive cost.

The editing distance between two trees corresponds to the minimum cost obtained in transforming one tree into another one with the help of these three operations. If this formalism is applied on a k-dimensional space modeled by a binary tree, it can be noticed that:

- the labels of tree nodes consists in the alphabet white, black, grey, also represented by,

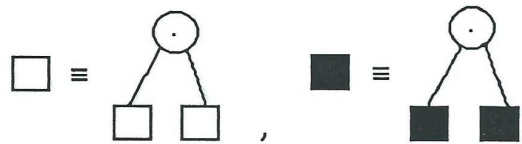

- the binary trees are complete : any node cannot be partially developed,
- the terminal nodes are equivalent to the trees :

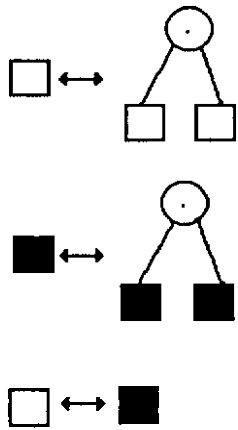

The set of all the parts belonging to all the spaces of any dimension is provided by applying the following grammar rules:

□ ↔ △(□, □)

■ ↔ △(■, ■)

□ ↔ ■

where derivation rules are bi-directional. So concerning binary trees, the inserting or the deleting of a node do not exist, only changing a label gets some truth.

Knowing that a node of a given color can be divided into two new nodes of the same color, the cost of a replacement made at depth $p+1$ in a tree should cost half less than at depth $p$, if the label replacement cost is 1 at the root of the space, it should be at level $p$ :

$$c_p(\square \leftrightarrow \blacksquare) = \frac{1}{2^p}$$

This value is also the hypervolume of the $2^k$-ant associated to a node at depth $p$ in a tree.

Knowing that the binary trees are sentences of infinite length and that it is consequently possible to directly compare them node by node, the minimal cost transform is unique, it is the one that changes node by node the labels of differently colored nodes between them along a parallel traversal of the two trees.

Concerning the replacement cost described above, the cost of this transform is equal to the mass of the exclusive or of the two modeled sets, that is also the Lebesgue measure of the difference of the two sets in another words the Hausdorff distance applied on sets modeled by $2^k$-trees:

$$d(A, B) = \mu(A \oplus B)$$

That is also the weighted extension of the Hamming distance applied on tree codes.

The topology induced by the $2^k$-trees is less thin than the metric topologies commonly used. In fact if the values 0 and 1/8 have got 1/8 as distance measure, the values 3/8 and 4/8 or 2/8 and 5/8 will get 7/8 as distance measure. It is only depending on the height of the common ancestor of these values in a tree, which is providing the property of ultrametricity to this distance.

The Hausdorff distance is an ultrametric distance that does not enable to measure the distance between two points of a space, but the distance between two parts of this same space.

# III – Generation of trees modeling multidimensional data sets

## III.1 - Representation of a vector

For building the tree of a multidimensional data set, it will be assumed that data will be available in the form of integer vectors of dimension k (cf. figure 5). Each vector element would have previously normalized inside the interval $[0, 2^r - 1]$, where $r$ will represent the maximum precision of handled data. So if $u \in [0, 2^{r-1}]^k$ is represented by a monodimensional table $\{u(i), i = 1, k\}$, each of its elements $u(i)$ will get as binary description:

$$u(i) = C_1 C_2 ... C_r \text{ where } C_j \in \{0,1\} \text{ for } j = 1, r$$

and will refer to the unsigned natural integer :

$$u(i) = C_1 \times 2^{r-1} + C_2 \times 2^{r-2} + ... C_r \times 2^0$$

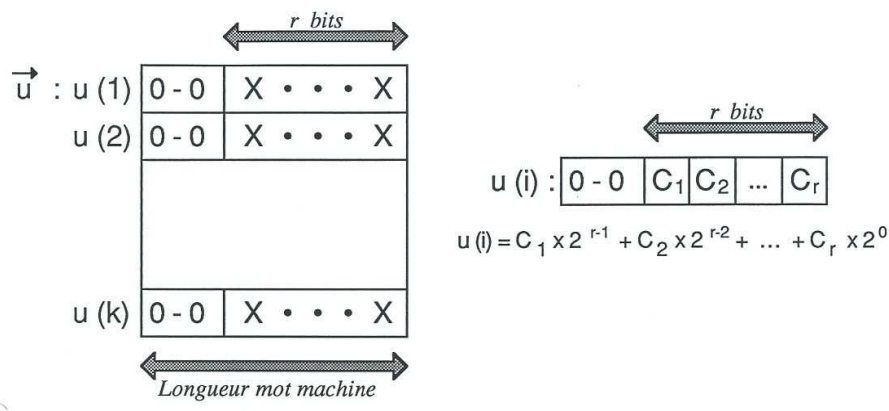

**Figure 5 : Format of a normalized vector**

To generate the binary tree representing the vector $\vec{u}$ in the discrete space of k dimensions

$\{0, 1, , 2^{r-1}\}^k$, also assimilated to $\left\{0, \dfrac{1}{2^r}, ..., \dfrac{2^r - 1}{2^r}\right\}^k$, is equal to :

- from the root, to analyze the bit $C_1$ of the first element $u(1)$ and to mark white the left son if $C_1$ is equal to 1, the right son in the contrary case ;
- when the bits $C_1$ have been exhausted, it must be come back to $u(1)$ and restarted with the following bit $C_2$ ;
- the process is iterated until having reached the resolution $r$.

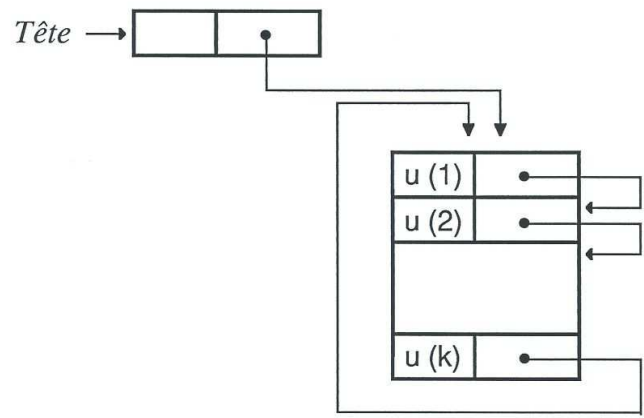

**Figure 6 : Circular linked list of a vector**

The generation is performed down to the maximum precision $k \times r$ in the space:

$$\left\{ 0, \frac{1}{2^{kr}}, ..., \frac{2^{kr}-1}{2^{kr}} \right\}$$

This formalism can be also applied to the normalized real numbers. The division by $2$ is used for extracting a binary representation of the vector elements. So the generation of the tree of a vector can be performed by the following recursive algorithm:

    depth <— dimension * precision

    root <— tree of vector (vector, 0, depth)

    /* Generation of the tree of a vector */

    FUNCTION tree of vector (vector, level, depth)

    BEGIN

      IF (level = depth) THEN RETURN (*tree* (*black*))

      ELSE DO

          side <- extraction of most significant bit and left shift of the present coordinate of the vector

          IF (side = *left*)

          THEN RETURN (*sub-trees union* (tree of vector (*rotation* (vector), level + 1, depth), *tree* (*white*))

          ELSE RETURN (*sub-trees union* (*tree* (*white*), tree of vector (*rotation* (vector), level + 1, depth))

END

        END

## III.2 - Generation of the tree of a set of vectors

By modifying the previous algorithm, it is possible to generate the tree modeling a set of vectors. Each vector represents a realization of the indicator function that has been previously described. That is to say that a vector $\vec{u} = (u_1, u_2, ..., u_k)$ will be an element of $S = \{\vec{u} / \delta(\vec{u}) = 1\}$.

The new proposed algorithm does not create the tree of a vector, but enriches an existing tree with the realizations of a set of vectors. Enriching this structure consists in generating the path or the part of the path, which is still described in the tree, when the circular list of a vector is analyzed.

The initialization of a tree generation is performed by creating a white colored tree. This procedure generates a tree with only two white nodes, which is the set $U = \{\forall \vec{u} / \delta(\vec{u}) = 0\}$ and which is representing the empty universe of any dimension and at any precision. The initialization implies that whatever is $\overline{S}$, any path belonging to $\overline{S}$ is already registered in the tree. So only the black paths that are still not described should be generated.

The merging of black symmetrical paths, on the return of recursive calls, enables to aggregate the uniformly colored nodes.

    root <- *tree* (*white*)

    depth <- dimension * precision

    CALL addition of a vector (root, vector, 0, depth)

    /* <u>Addition of a vector to a tree</u> */

    PROCEDURE addition of a vector (root, vector, level, depth)

    BEGIN

        IF (level = depth) ALORS *blackening* (root)

        ELSE DO

            side <- extraction of most significant bit and left shift of the current coordinate of the vector

            IF (*terminal* (root)) THEN *fission* (root)

            CALL addition of a vector (*child* (root, side), *rotation* (vector), level + 1, depth)

            *merge* (root)

        END

END

This procedure of generation of the tree of a multidimensional data set by enriching a structure shows different advantages:

- it enables to take in account overcrowded data sets ;
- it is compatible with ordered or not data flows.

## *III.3 – Tree access operators and associated algorithmic*

The algorithmic applied on $2^k$-trees, which is followed, is built on the recursive calculus. It takes in favor depth-first traversal of tree-like structures. It follows the definition of a tree proposed by KNUTH who was understanding it as the recursive concatenation of sub-trees. So any recursive operator will fit to the root of a tree, as well as to each of its nodes.

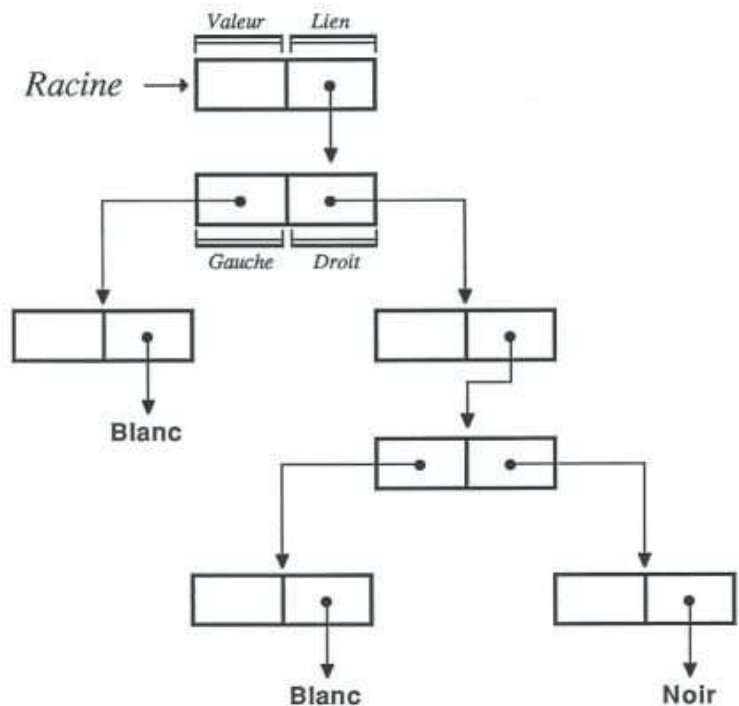

**Figure 7 : Binary tree structure**

An empty tree will have no meaning. This notion is replaced by:

- a white node, as representing an empty set of any dimension and at any precision ;
- a black node, as representing the entire space (full set) of any dimension and at any precision.

That is meaning according to the fact that a node is terminal or not. The implementation, which has been developed for modeling a tree node, relies on the use of:

- two double-words for a non terminal node ;

– a single double-word for a terminal node.

The first double-word is holding the addresses of the left and right sons of a node (the sub-trees). Concerning a terminal node, this second word is useless because the white and black addresses are auto-referring in the implemented addressing system. Knowing that any node in a tree is the root of a new tree, it will be applied the word of root for appointing the address of a node in a tree.

For processing trees, it will be used depth-first traversals. So, the provision of a tree according to whichever used operator can be performed with only two complementary ways:

- by node fission, if the generation is performed during the descent of recursive calls ;
- by node merge, if the generation is performed during the return of recursive calls.

A node deletion in a tree does not exist as it. It is replaced by the removal of the children sub-trees and the setting at terminal state of the father node.

In addition, so as to minimize the extension of a tree, a merging operator enables to transform a tree in which the two sons are iso-colored into a terminal node of the same color.

The it is possible to make an analogy between classical list operators and those that can apply on binary trees, this one is shown in the table underneath.

| List operators | Tree operators |
| --- | --- |
| Create a list | Create a tree of a given color (white/black) |
| Is empty a list? | Is terminal a node ? |
| Next in the list | Child of a given side (left/right) |
| Insert an element in a list | Fission of a node into two others or union of two sub-trees |
| Delete an element in a list | Merge a non terminal node |
| Delete a list | Destruction of a tree |

**Table 1 : Correspondence between list and tree operators**

## *III.4 – Boolean operations on trees*

A lot of algorithms implementing Boolean operations on 4-trees or 8-trees have been published. The Boolean operations on $2^k$-trees are performed according to the rules of the set theory.

Let be two k-dimensional sets $S_1$ and $S_2$, represented by the binary trees, $tree(S_1)$ and $tree(S_2)$. The Boolean operations «and», «or», «exclusive or» and «not» are defined according to the following manner:

- $tree(S_1)$ and $tree(S_2) \rightarrow tree(S_3) \:/\: S_3 = S_1 \cap S_2$
- $tree(S_1)$ or $tree(S_2) \rightarrow tree(S_3) \:/\: S_3 = S_1 \cup S_2$
- $tree(S_1)$ xor $tree(S_2) \rightarrow tree(S_3)) \:/\: S_3 = S_1 \oplus S_2$
- not $tree(S_1) \rightarrow tree(S_3) \:/\: S_3 = \overline{S_1}$

When the sets are representing objects of k dimensions, the Boolean operation "and" will perform the intersection of the hypervolumes describing the objects, the operation "or" their union.

As terminal nodes of a binary tree have been implemented using auto-referring values, no troubles will appear when paths of unequal lengths will be compared between two trees. This feature enables to compare two trees built with two different precisions and to produce a new tree at a precision that can be once more different from its operands. It makes available in addition to provide an assertion operator that builds a copy of the operand tree but at a different precision from that used for its generation.

When a non terminal node is encountered at the maximum precision of calculation, it has been chosen to process it as it was a black node. So the operators performed at variable precision, are doing it according to the uppermost hull, that is meaning that the resulting trees will be nested into each another one, while the precision will decrease. For instance, the union of two trees modeling two objects in a space of any dimension follows the next algorithm:

```
depth <- dimension * precision

root <— union (root1, root2, 0, depth)

/* Union of two binary trees */

FUNCTION union (root1, root2, level, depth)

BEGIN

    IF ((NOT terminal (root1)) OR (NOT terminal (root2)))
    AND (level ≠ depth)) THEN DO

        /*descent in depth of the two trees*/

        root <- sub-trees union (

        union (left son(root1), left son(root2), level+1, depth),

        union (right son(root1), right son(root2), level+1, depth))

    END
```

```
        ELSE DO

        /*union of the two reached nodes*/

        IF ((white(root1)) AND (white(root2)))

            THEN RETURN (tree (white))

            ELSE RETURN (tree (black))

        END

        /*merge of children nodes when going up in the tree*/

        merge (root)

        RETURN (root)

    END
```

## III.5 – Inductive limit computations

In order to implement a procedure that will preserve the topological organization and the ability of directly comparing trees between themselves, an initial topological structuring must be defined over the spaces including any data of $R^k$.

At first, it must be noticed that using a tree for modeling a data set of $[0,1]^k$, it leads to give it a structure of Borel algebra. This structure is built on the atoms made from the $2^k$-ants resulting from the meshing of $[0,1]^k$ down to the precision $r$.

For each point of $[0,1]^k$, it can be combined a fundamental neighborhood system made from the atom which is including it and the set of $2^k$-ants linked to the fatherly nodes of the tree branch which allows to reach this node from the tree root: it will be got in such a way a series of nested parts whose union will make up the unitary hypercube.

They are these fundamental neighborhood systems that, by symmetries analysis that they are sharing between themselves, will enable to deduce from the modeled sets the properties that they are sharing under the $d_1$ and $d_\infty$ metric topologies and to transform them while respecting these ones.

Each $2^k$-ant of $[0,1]^k$ whose precision is included between $1$ and $r$, that is belonging to a branch of the initial tree, makes up in its turn a Borel algebra down to an intermediate precision and is included in the original Borel algebra and shares the same topology restricted to this sub-set.

In a similar manner, any algebra built as a union of an appropriate number of translated $[0,1]^k$ so as this one appears as homothetic to $[0,1]^k$ will preserve once more the initial topology of $[0,1]^k$, deduced from its neighborhood system.

So it can be propose an approach for building trees such as $R^k$ is the inductive limit of the algebras homothetic to the unitary hypercube and such as the topologies of these algebras remain compliant with those of their sub-algebras, that is meaning that a sub-algebra will always appear as a branch of the tree of an algebra.

The implemented building method has been adapted from the tree generation by enrichment. The approach is the following one, when a vector is added to an existing tree:

- if the vector is not included in the initial space then the new bounds of the right space holding the vector are computed an the tree is extended up to these new bounds ;
- the vector is normalized according to these bounds, then added to the data already present in the tree.

Computing new bounds, it is to determine the coordinates of the hypercube homothetic of a power of 2 of the previous hypercube, which is holding both this hypercube as well as the new data vector. These coordinates are valued at:

- in integer numbers :
$$x_{\min,i} = d\left(\min\{x_{\min,i}, x_i\}/d\right)$$
$$x_{\max,i} = d\left(\min\{x_{\max,i}, x_i\}/d + 1\right) - 1$$
- in real numbers :
$$x_{\min,i} = d \cdot \lfloor \min\{x_{\min,i}, x_i\}/d \rfloor$$
$$x_{\max,i} = d \cdot \lceil \max\{x_{\max,i}, x_i\}/d \rceil$$
- where $d = 2^{\log_2(x_{\max,i} - x_{\min,i})}$ for $i \in \{1,...,k\}$
- where $\{x_i\}_{i \in \{1,...,k\}}$ is the data vector, and $\{x_{\min,i}\}, \{x_{\max,i}\}$ the ends of the first diagonal of the hypercube ;
- and where $\lfloor \; \rfloor$ et $\lceil \; \rceil$ are the floor and ceiling operators for converting a floating point number into an integer one.

These computations are implemented by an iterative procedure which is converging towards the satisfying of these equalities.

The generation of a tree in inductive limit needs to keep always some complementary information: the bounds of the space in which has been built the $2^k$-tree. These bounds are the vertices of the hypercube of the tree decomposition, which is also the referring hypercube of the set modeled by the tree.

Until now the $2^k$-trees have always been built in the unitary hypercube of the space at k dimensions.

Building in inductive limit enables to build trees in any translated of a homothetic copy of the unitary hypercube. For actually comparing two trees generated into two different referring hypercubes, it is necessary to know these ones.

When the $2^k$-trees are brought together with their referring hypercubes, then it can be performed Boolean operations on them, even if their hypercubes are not equal. By comparison of the operand referring hypercubes, it can be know which hypercube is including them and how to extend the operands to this new hypercube. The two trees having been extended up to this new space, the Boolean operation is applied and provides a resulting tree which referring hypercube is the hypercube computed from those belonging to the operands.

# IV – Geometric transforms

## IV.1 – Tree of a polytope

We will focus on a specific class of polytopes (hyper-polyhedrons), the homographic transforms of the unitary hypercube. These polytopes have:

- $2^k$ vertices (confused or not),
- $2k$ faces (parallel for the linear transformed image of the unitary hypercube).

They can be represented by:

- the list of the coordinate vectors of their vertices ;
- the list of their faces divided into two sub-lists, the lower faces, the upper faces (dual representation).

Concerning this last representation, the polytope is defined as the intersection of the right half-spaces of the lower faces with the left half-spaces of the upper faces (cf. figure 8) :

- if $P_{\min 1}, P_{\min 2}, ..., P_{\min k}$ are the equations of the lower faces
- and $P_{\max 1}, P_{\max 2}, ..., P_{\max k}$ those of the upper faces,

the polytope is then the set :

$$\{u \,/\, {}^tP_{\min 1} \cdot u \geq 0\} \cap \{u \,/\, {}^tP_{\min 2} \cdot u \geq 0\} \cap ... \cap \{u \,/\, {}^tP_{\max k} \cdot u \leq 0\}$$

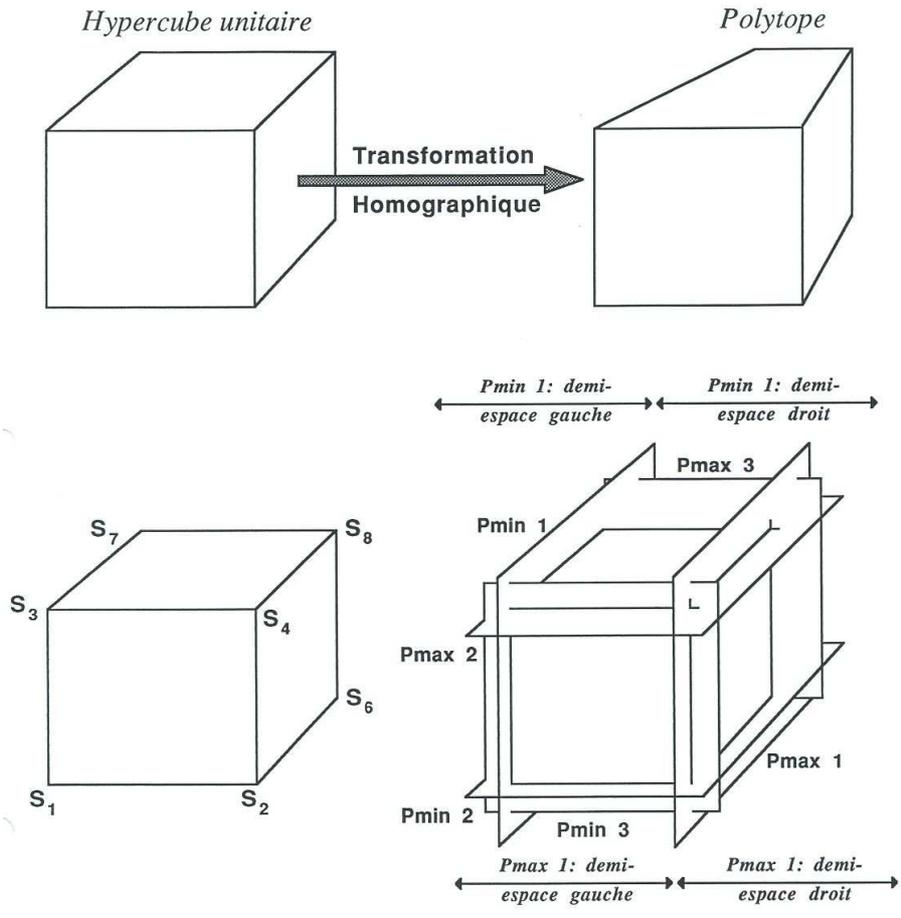

**Figure 8 : Representation of a polytope**

In this way, it can be obtained the particular transformed images (cf. figure 9):

- the translated image of the cube of the origin ;
- by applying rotations centered at the origin of the universe (generalized rotation) or at the middle of the universe (generalized rotation composed with a translation and its inverse) ;
- the equivalent of a hyperplane of the space by applying an anamorphosis of ratio $2^{-r}$ according to the desired axis, $r$ being the precision of computation (cf. figure 9) ;
- a pyramid by a perspective of which the center will be its top vertex ;
- a pyramid of view of a perspective, by truncation of front and back view-planes.

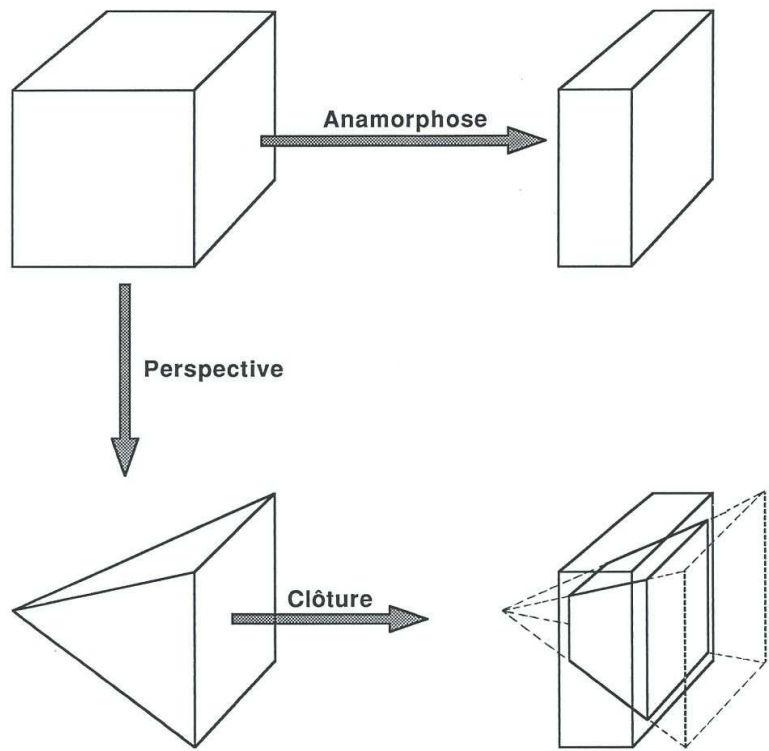

**Figure 9 : Particular transformed figures of a hypercube**

The building of the tree of a polytope is based on the recursive dividing of the unitary hypercube and the interiority test of the divided parts with the polytope.

The recursive dividing of a parallelotope (the hypercube) half by half can be extended to its homographic transformed images (the polytopes). This dividing takes in account of the particular order in which are put:

- the vertices of the polytope,
- the hyperplanes of the polytope.

The vertices are divided by packets of $2^{k-1}$, $2^{k-2}$, ... , $2$ vertices according to the same arrangements as those applied in a fast Fourier transform.

For performing the dividing of a polytope according to its first direction, the vertices will be determined by computing for all the intermediate vertices the averages between the first $2^{k-1}$ vertices and the following $2^{k-1}$ ones.

Then, for dividing according to the following direction, among the first $2^{k-1}$ vertices, part of the new vertices will be obtained by computing the averages between the first $2^{k-2}$ vertices and the following $2^{k-2}$ ones, and the other part by performing the same operation over the remaining $2^{k-1}$ vertices.

And so on, until the dimension k where the new vertices are the direct averages of the points two by two.

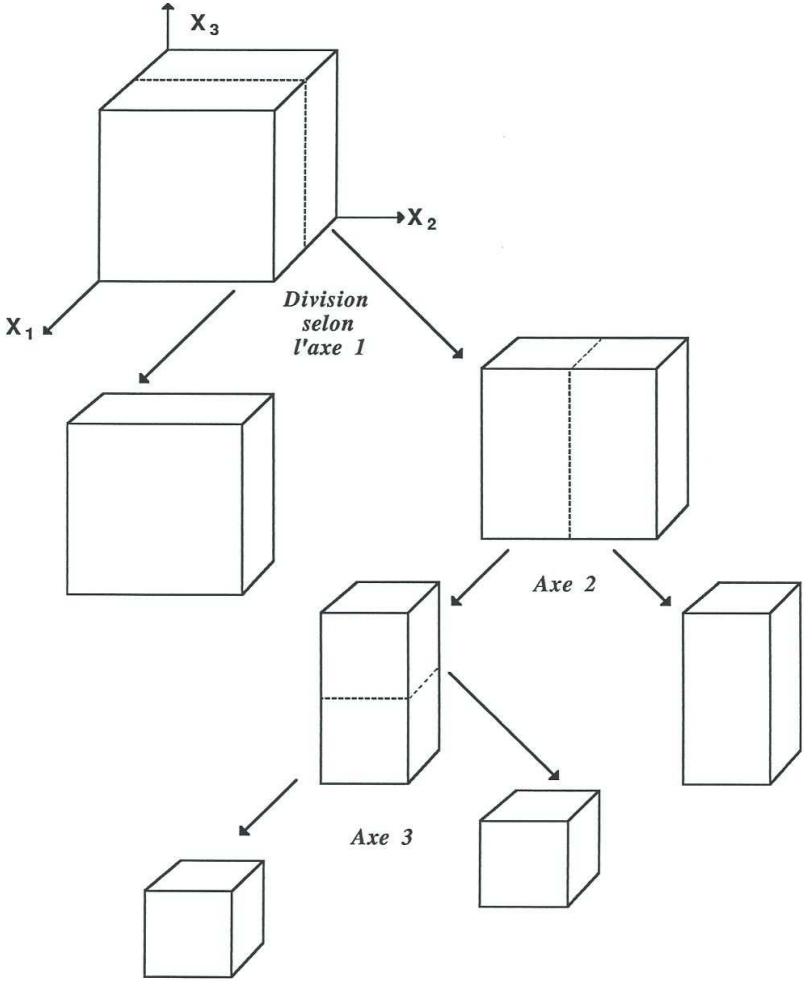

**Figure 10 : Recursive dividing of a cube**

Concerning the planes, these ones are enriched by the median planes of the lower and upper planes according each division direction (cf. figure 11).

For a polytope defined by its faces, it is simpler, because the median plane has for equation the half-sum of the equations of the lower and upper planes along the current dimension.

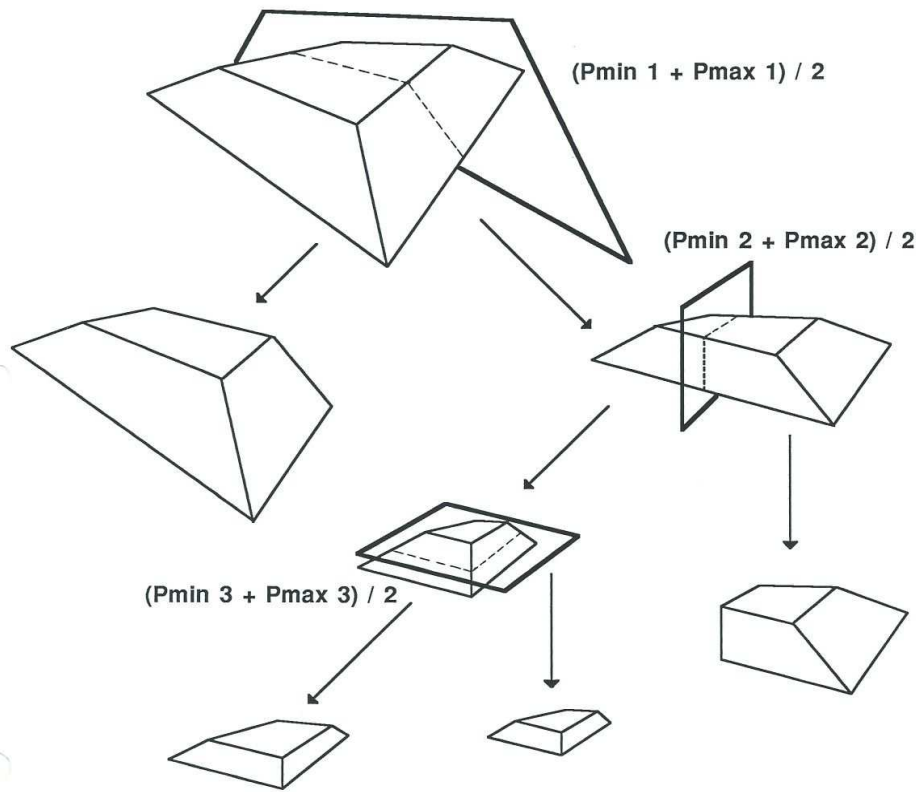

**Figure 11 : Division of a polytope**

Having at our disposal two dual representations of a polytope, here is how to evaluate the intersection of a $2^k$-ant and a polytope.

A polytope is a convex set that owns the following properties:

- each point internal to a polytope is a linear positive combination of its vertices (barycentric coordinates) ;
- a polytope is entirely located on the same side of each hyperplane defining its faces : that is to say inside the positive half-spaces of its lower planes and the negative half-spaces of its upper planes

Let be two polytopes to compare, if all the vertices of a polytope are located in one of the half-spaces external to the other polytope, it will be the same for any convex combination of these points, then the two polytopes do not get any intersection.

If all the vertices of a polytope are located inside all the inner half-spaces defined by the faces of the other polytope, it will be the same for any convex combination, then it is inside this polytope.

Finally, if the vertices of a polytope are located on both sides of one of the faces of the other polytope, there is an intersection but without any inclusion of one into the other one.

By recursively dividing the unitary hypercube and comparing the result with the initial polytope, it can be directly generated the tree of this polytope while coloring in black the inclusions, in grey the intersections to be developed and in white the lacks of intersection.

## IV.2 – Homographic transformation of a $2^k$-tree

Analytically, a hyperplane is the set of points:

$$\{u \in E \,/\, {}^t v \cdot u = 0, v \in E^*\}$$

i.e., by decomposing $u$ in its reference system and $v$ in its dual basis:

$$\left\{u \in E, \sum_{i=1,k} v_i \cdot u_i = 0, v \in E^*\right\}$$

One of the main interests of this analytic representation is the following one: if a bijective map $f$ is applied on a set of points $E$, it is equivalent to perform the inverse map $f^{-1}$ on $E^*$.

So the homographic transformed image of a polytope gets:

- as vertices, the direct transformed images of its vertices;
- as faces, the inverse transformed images of the parametric expressions of its faces.

In affine coordinates, the displacements that can be applied on a space of k dimensions are the translations and the rotations:

$$X' = RX + T \quad \text{in affine coordinates,}$$

$$\begin{bmatrix} X' \\ 1 \end{bmatrix} = \begin{bmatrix} R & T \\ 0 & 1 \end{bmatrix} \begin{bmatrix} X \\ 1 \end{bmatrix} \quad \text{in homogenous coordinates.}$$

where $R^{-1} = {}^T R$ et $T^{-1} = -T$

They are the movements that can be applied on rigid bodies. Completed with the homotheties, they constitute the group of similarities:

$$X' = HRX + T, \text{ où } H = \lambda I_k$$

If the homotheties are extended up to the anamorphoses, it is obtained the linear positive group of $E$, $GL(E,+)$:

$$X' = ARX + T, \text{ where } A \text{ is a positive diagonal matrix.}$$

If the axial symmetries are added, it is provided the linear group of $E$, $GL(E)$, set of the linear maps of $E$:

$$X' = ARX + T, \text{ where } S^2 = I_k, \text{ i.e. } S^{-1} = S$$

If the space is described in homogenous coordinates, the linear group, completed with perspectives, constitutes the projective linear group $PGL(E)$:

$$\begin{bmatrix} X' \\ W' \end{bmatrix} = \begin{bmatrix} SAR & T \\ {}^T P & 1 \end{bmatrix} \begin{bmatrix} X \\ W \end{bmatrix}, \text{ where } P^{-1} = -P$$

It gathers all the maps en in homogenous coordinates that can be applied on $E$, the maps being geometrically together equivalent by a multiplicative factor. They are homographies.

For a space described in affine coordinates, the homographic transformations are not linear. They are needing at the end of a transformation to normalize the homogenous coordinates with the help of the weight of the (k+1)-th coordinate in order to able to come back to affine coordinates.

The process of recursive dividing of a polytope can be also applied on the calculation of the homographic transformation of a tree. Actually, the middle of a line segment is in harmonic division with the ends of this line segment and the point at the infinite according to the direction of the line segment (their cross-ratio is valued at $-1$). The cross-ratio of four points remains unchanged by any homography.

By duality, two hyperplanes, their median hyperplane and the hyperplane at the infinite constitute a harmonic bundle.

Then there will be an equivalence between the homographic images of the decompositions of an hypercube and the recursive decompositions of the homographic image of the same hypercube.

To avoid the computation of the homographic transformed images of $2^k$-ants inside the initial space, it can be noticed that it will be got a same tree as the tree of the transformed set by decomposing the initial set into the inverse image of the unitary hypercube of the image space. Actually, the vertices of a regular division of the unitary hypercube of the image space and their inverse images will be in bijective correspondence for this transformation.

For computing the image tree resulting from this transform, it is then equal to decompose the transformed set inside the unitary cube as well as to decompose the initial set inside the inverse image of this same cube. For instance, the figure 12 shows this fact for a rotation of a given angle: the decomposition of the transformed image by the given rotation is bijective with the decomposition of the initial set in the inverse image of the unitary hypercube (bijection enlightened by the hatched part in a same quadrant of decomposition).

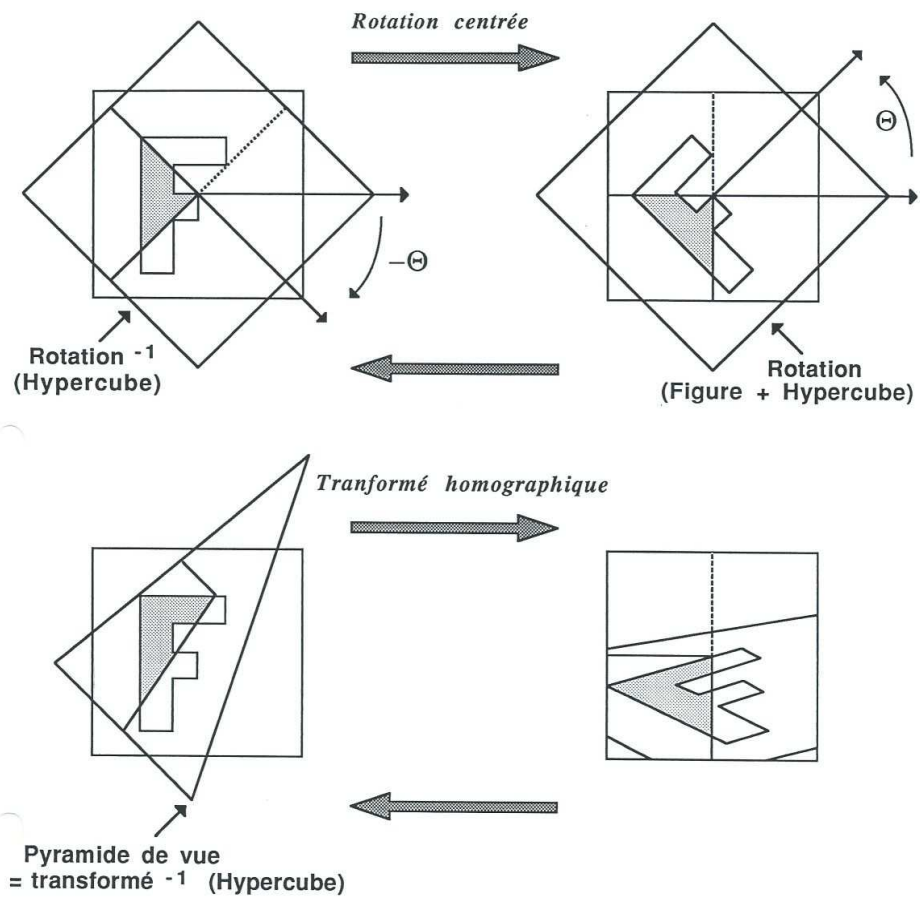

**Figure 12 : Calculation of the homographic image of a set**

It can be then noticed that the only information of the inverse image of the unitary hypercube is enough for defining the homographic transformation. This one is fully described by:

- the list of the inverse transformed vertices of the unitary hypercube,
- the list of the direct images of the lower and upper faces of the same hypercube.

When this reasoning is applied to a perspective transform, it can be seen that the inverse image of the unitary hypercube is then the pyramid of view associated to this transformation (cf. fig. 12). When it is not truncated, it is a prism of which the top is the center of perspective. The closing of the view is performed by truncation of this pyramid.

For assessing the node color of the image tree, it must be compared the intersection of the blocks of the original tree with those provided from the regular decomposition of the transformation polytope, according to the principle implemented for the assessment of the intersection of two convex polytopes. This approach remains available because the harmonicity of recursive divisions preserves the block convexity

This building method has got the advantage of being indifferent to the lacks of covering between the set to be transformed and the grid of decomposition.

Finally, the calculation of the homographic image starts with the building of the tree of the transformed polytope. The building is driven by the search of common black nodes between the set to be transformed and the tree of the transformed polytope, which is restricting the calculation of the transformed image to the only nodes concerned by the transformation inside a tree of any precision.

# V - Segmentation

## V.1 – Adjacencies search

The concept of neighborhood in a metric space is relying on the use of a distance in this space. The most commonly used distances are:

- $d_\infty(X,Y) = \max_{i=1,k} |x_i - y_i|$,
- $d_1(X,Y) = \sum_{i=1,k} |x_i - y_i|$,
- $d_2(X,Y) = \left( \sum_{i=1,k} |x_i - y_i| \right)^2$, Euclidean distance.

Inside a meshed metric space, two points $X$ and $Y$ will be adjacent or neighbors, if they are distant of one resolution unit of the space. That is to say that they must satisfy to the relation:

$$X \, \Re_d \, Y \,:\, X \neq Y \text{ and } d(X,Y) \leq 1 \quad \text{in } \{0,1,...,2^{r-1}\}^k,$$

or again :

$$X \, \Re_d \, Y \,:\, X \neq Y \text{ and } d(X,Y) \leq \frac{1}{2^r} \quad \text{in } \left\{ 0, \frac{1}{2^r}, ..., \frac{2^r - 1}{2^r} \right\}^k.$$

So the set of neighbors of $X$ in $\{0,1,...,2^{r-1}\}^k$ will be the unitary ball:

$$B_d(X,1) = \left\{ Y \in \{0,1,...,2^r - 1\}^k \,/\, Y \neq X \text{ et } d(X,Y) \leq 1 \right\},$$

And in $\left\{ 0, \frac{1}{2^r}, ..., \frac{2^r - 1}{2^r} \right\}^k$, it will be the ball of highest resolution $B_d\left( X, \frac{1}{2^r} \right)$.

The adjacency degree is depending on the dimension, the distance and the mesh defining the support space. Concerning squared meshes (cf. figure 13), there are:

- 4 $d_1$-neighbors in 2 dimensions, 6 $d_1$- neighbors in 3 dimensions ;
- 8 $d_\infty$- neighbors in 2 dimensions, 26 $d_\infty$- neighbors in 3 dimensions.

In one dimension, there are only 2 $d_1$- or $d_\infty$- neighbors.

The Euclidean distance, is not directly reachable with a meshed space managed by a $2^k$-tree.

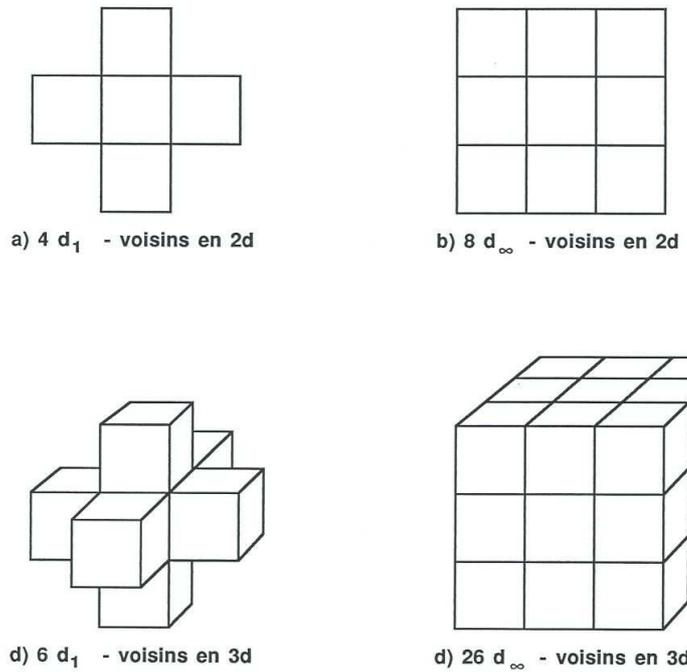

**Figure 13 : Adjacency degree according to the used metric space**

Several works have been led on the search of $d_1$-adjacencies. They are all based on the search of a common ancestor of two distinct points in a $2^k$-tree. The applied method is the following: for each terminal node, all their ancestors are examined, for a given ancestor under analysis it is enough to generate a mirror walk according to the space axes comparatively to this one that has enabled to reach the ancestor, in order to look for the candidates to an adjacency.

The figure 14, taken back from ([SAMET 82a]), is showing its principle. This approach is well adapted to tree traversal in post-order.

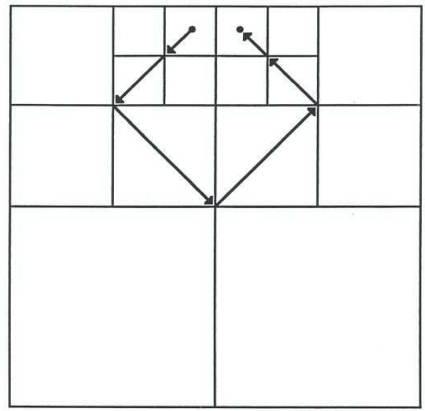

**Figure 14 : Looking for a common ancestor in a quaternary tree**

On $2^k$-trees visited in a descending way, and more especially in depth first, it is proposed to follow the opposite way to this approach: for each non terminal node, it is proposed to develop all the possible adjacencies and to check them gradually during the tree descent.

As it is shown by JACKINS et TANIMOTO ([JACKINS 83]), the $d_1$-adjacencies are using only a single symmetry plane (or hyperplane) per adjacency between $2^k$-ants.

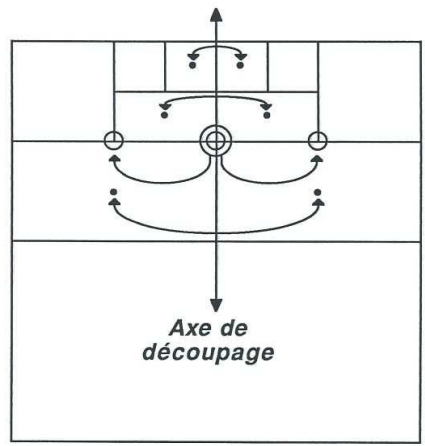

**Figure 15 : Looking for $d_1$-adjacencies in a 4-tree represented by a binary tree**

Let us assume that the binary tree modeling a quaternary tree is visited according to a descending way (cf. figure 15).

The two sons (represented by a single circle) of a non terminal node (represented by a double circle) are $d_1$-adjacent according to the dividing axis of the non terminal node

These two sons have got grandsons. Among these ones, are adjacent:

- on the one hand, the grandsons coming from the division along an orthogonal direction to the father dividing axis and this performed side by side (left grandson with left grandson, and with the same manner for right grandsons).
- on the other hand, the left and right grandsons coming from the left and right sons, along a parallel direction to the father dividing axis.

So it will be obtained all the $d_1$-adjacencies of a non terminal node in a binary tree modeling a universe of dimension k.

The parallel directions to the initial dividing direction are those that are creating the mirror effect used by the search methods of a common ancestor: they are orthogonal to the symmetry plane providing the $d_1$-adjacency.

Only SAMET has been interested in looking for $d_\infty$-adjacencies and this study has been performed on quaternary trees.

In a universe of dimension k, the $d_\infty$-adjacencies are including the $d_1$-adjacencies, which is meaning those produced by the adjacencies around a dividing axis among k possible ones. They can be completed with the adjacencies created by the connections generated by two or more than two axes among the k possible ones, that is to say by the combination of two or more than two symmetry planes or hyperplanes.

So, as it is shown on figure 16.a) in two dimensions:

- after two dividing steps, the furthest nodes coming from the sons are adjacent;
- all the grandsons, coming from a further division according to each of the two axes, are adjacent, right and left grandsons coming from the left and right sons (mirror effect).

In three dimensions (cf. figure 16. b) ; the same situation occurs:

- for the two diving axes (link 1 with 2), the third one does not play any role,
- for the three axes (link 1 with 3),
- and for the first and the last dimensions, even if it is not displayed on the figure.

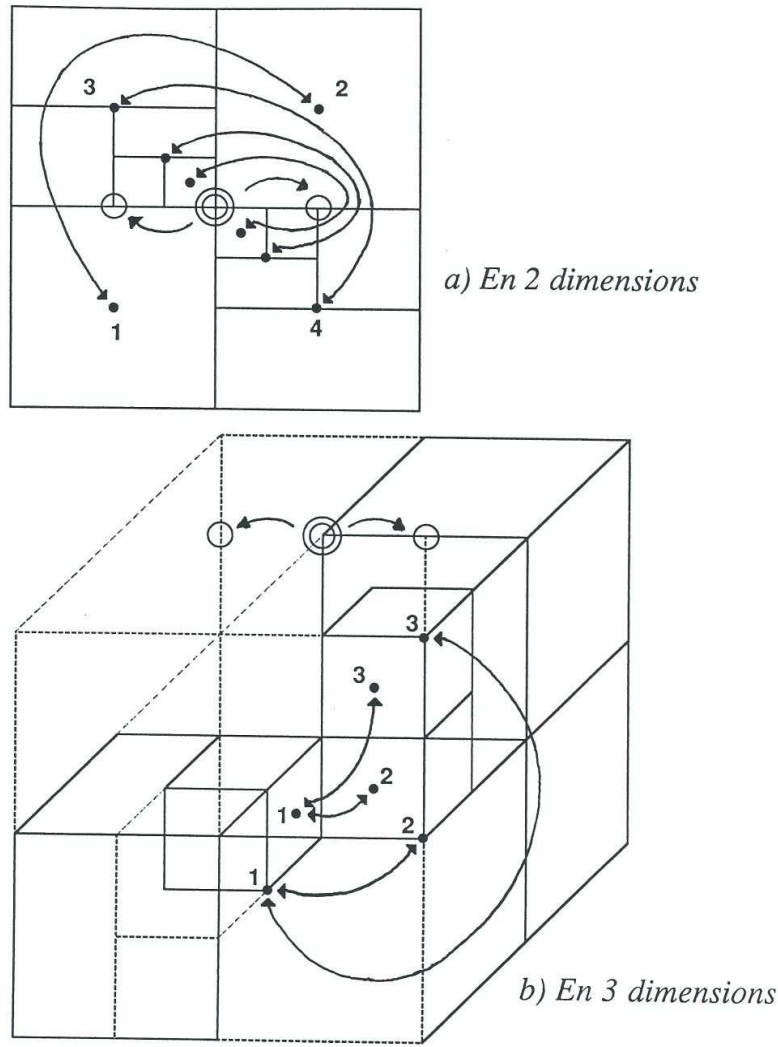

**Figure 16 : Looking for $d_\infty$ -adjacencies in a binary tree**

In k dimensions, the $d_\infty$ -adjacencies are generated by the symmetries:

- around each of the k possible axes ($d_1$ -adjacencies),
- around 2 among k axes, for all the $C_k^2$ possible combinations,
- and so on, up to k axes.

Knowing that the symmetries generated by j axes among k ones are as many as the number of dividing possibilities left free among the k-j remaining axes, then there are $2^{k-j}$ possible symmetries inside a $2^k$ -tree for the sons of a single node.

At the opposite, at a given resolution level, the number of neighbors that a single node may have is deduced from the number of symmetries generated by the divided of these same ancestors: for j axes among k possible ones, the node will get $2^j$ neighbors.

So at a given resolution level inside a $2^k$ -tree, a node could get:

- $2^1 C_k^1 = 2k$      $d_1$-neighbors,
- $\sum_{j=1}^{k} 2^j C_k^j = 3^k - 1$      $d_\infty$-neighbors.

To take benefit of the mirror effect, it is necessary to store the initial positions of the grandsons so as to implement correctly the mirror effect, which enables to link the grandsons from opposite sides to initial sides.

In a space of k dimensions, are registered in a circular vector, the initial configurations of the so-generated symmetries, with the help of an indicator:

- N : neutral, for a dimension not taken in account,
- A : anti-symmetric, for the right and left sons,
- S : symmetric, for the left and right sons.

The $d_\infty$-adjacencies are provided by the combination of k symmetries at more taken among the k directions of the space. The $d_\infty$-adjacencies are generated by the intersections of all the dividing hyperplanes of the mesh, that is to say for a different direction from the initial dividing direction, by all the directions orthogonal to this first direction, that can come across at any level in the sub-tree.

The analysis of a sub-tree in order to detect all the $d_\infty$-adjacencies is then formalized by the following recursive procedure:

    CALL search adjacencies (*left son*(root), *right son*(root), *(S, N, …, N)*)

/* <u>Looking for adjacencies</u> */

PROCEDURE search adjacencies (node 1, node 2, (…, …, …))

BEGIN

    indicator <- first list element (…, …, …)

    IF (*non terminal* (nodes)) THEN DO

        *copy* (…, …, …) *for recall*

        IF (indicator=*N*) THEN DO

            search adjacencies (*left son* (node 1), *left son* (node 2), *rotation* (N, …, …))

            search adjacencies (*right son*(node1), *right son*(node2), *rotation*(N, …, …))

            search adjacencies (*right son*(node 1), *left son*(node 2), *rotation* (A, …, …))

            search adjacencies (*left son*(node 1), *right son*(node 2), *rotation* (S, …, …))

        END

IF (indicator=*S*) THEN

            search adjacencies (*right son*(node 1), *left son*(node 2), *rotation*(*S*, ..., ...))

        IF (indicator=*A*) THEN

            search adjacencies (*left son*(node 1), *right son*(node 2), *rotation*(*A*, ..., ...))

        *restore* stored vector

    END

    ELSE *store* the adjacencies of terminal nodes

END

It can be come back to the $d_1$-adjacencies by deleting the combinations of symmetry axes:

CALL search adjacencies (*left son*(root), *right son*(root), *(S, N, ..., N)*)

/* <u>Looking for adjacencies</u> */

PROCEDURE search adjacencies (node 1, node 2, (..., ..., ...))

BEGIN

    indicator < - first list element (..., ..., ...)

    IF (*non terminal* (nodes)) THEN DO

        *copy* (..., ..., ...) *for recall*

        IF (indicator=*N*) THEN DO

            search adjacencies (*left son* (node 1), *left son* (node 2), *rotation* (*N*, ..., ...))

            search adjacencies (*right son*(node1), *right son*(node2), *rotation*(*N*, ..., ...))

        END

        IF (indicator=*S*) THEN

            search adjacencies (*right son*(node 1), *left son*(node 2), *rotation*(*S*, ..., ...))

        *restore* stored vector

    END

    ELSE *store* the adjacencies of terminal nodes

END

## V.2 - Labeling of connected components

Two adjacent points $X, Y$, according to an unspecified distance, will be 1-connected. If there is a path of n-1 adjacent points enabling to join $X$ to $Y$, then $X$ and $Y$ will be n-connected.

A set of points will be a connected set, if whatever are two points $X$ and $Y$ belonging to this set, there is a path of an unknown length enabling to join $X$ to $Y$ in this set. In other words, a set $V$ will be connected if:

$$\forall X, Y \in V, X \, \Re_c Y \Leftrightarrow \text{either } X = Y,$$

$$\text{or } \exists \{Z_i\}_{i=1,n} \subset V / X \, \Re_c Z_1, \, Z_i \, \Re_c Z_{i+1}, i = 1, n-1, \, Z_n \, \Re_c Y.$$

The relation of connectivity $\Re_c$ is an equivalence relation.

Let be $V$, some set, a partition of $V$ is a decomposition of $V$ into n separate sub-sets $V_i$, as their union reconstitutes $V$:

- $\forall i, j / i \neq j \; V_i \cap V_j \neq \emptyset$;
- $V = \bigcup_{i=1,n} V_i$.

A predicate $P$ will a Boolean function over the set of the parts of some set $V$:

$$P : P(V_i) \in \{true, false\} \text{ and } V_i \subset V$$

A segmentation of $V$ according to the predicate $P$ is a partition $\{V_i\}_{i=1,n}$ of $V$ such as:

- $\forall i \in \{1, n\} \; P(V_i) = true$,
- and $\forall i, j \in \{1, n\}, i \neq j \; P(V_i \cup V_j) = false$

Let be $\Re$ an equivalence relation on $V$, an equivalence class of $\Re$ over the set $V$ will be a sub-set $V_\Re$ of $V$ such as:

- $\forall X, Y \in V_\Re \; V : X \, \Re \, Y$

The set of the equivalence classes of a set $V$ for a relation $\Re$ constitutes a partition of $V$.

The equivalence classes for the connectivity relation $\Re_c$ will be named connected components. Knowing that a multidimensional discrete space can be represented by its indicator function $\delta : \{V\} \to \{0,1\}$, it will be then provided a segmentation of $V$ by the predicate $P$ of iso-coloring:

$P(V) = true$ if $\forall X, Y \in V, \delta(X) = \delta(Y)$ over the set of connected components of $V$.

The adjacency analysis having been performed on the binary tree modeling a space of k dimensions, it is now possible to discover the connected components of this space and to label each terminal node with the label of the component to which it is belonging.

Two algorithms have been designed for implementing a recursive labeling over $2^k$-trees. There is the one developed by SAMET ([SAMET 81a]) which is relying on an algorithm described by ([KNUTH 73]), and the other one developed by GARGANTINI ([GARGANTINI 82c])

The SAMET algorithm is based on the notion of adjacency tree.

The method followed by GARGANTINI is easier to deal with but not easily parallelisable. It relies on the numbering of all the connected points belonging to a newly detected component:

- when a new component is detected, it is loaded up with the point that has enabling its discover and all its neighbors, and these ones are stored in a queue for their further processing;
- when one point has been examined, it is checked if the queue is not empty, if it is true, the next point to be examined is caught from the queue before looking for a new starting point in the tree, in order to exhaust the former detected component).

At last, the tree being labeled, it is then possible to compute the list of the segment-trees, that is to extract from the labeled tree, the trees of all the connected components and to link them into a single list : the forest of the segment-trees.

# VI – Attribute calculus

## VI.1 – Generalized moments and Eigen trees

After having segmented a scene and separated all the components which is composing it, one way for classifying the components comparatively to each other or to other ones previously analyzed, is to perform measurements over these components. It is also speaking about attribute calculus since it is enriching the components with numerical attributes. The measures the most commonly computed over objects are the generalized moments.

The generalized moments have already been applied on quaternary trees ([SHNEIER 81b], [RANADE 82]). Compared to other measures, the generalized moments show several different advantages:

- they are integral measures, which is meaning applied to regions and not boundaries, therefore not much sensitive to digital noise ;
- the moments are measures mutually independent from each others ;
- they enable to define measures insensitive to some geometrical transformations (the similarities).

This last property enables to build a representation of component trees that is invariant to these transforms [JAIN 82], [AGGARWAL 84]). It can then be spoken about normalized trees, it will be rather spoken about Eigen trees given that these ones are designed for showing objects in their Eigen reference frame and that it is existing other normalized forms for describing trees.

In a space of k dimensions, the generalized moments are the following measures:

$$M_{(objet)}\left(X_1^{n_1}, X_2^{n_2}, ..., X_k^{n_k}\right) = \int_{X_1} \int_{X_2} ... \int_{X_k (X_1, X_2, ... X_k) \in objet} X_1^{n_1} X_2^{n_2} ... X_k^{n_k} dX_k ... dX_2 dX_1$$

where :   $n_i \geq 0, \forall i \in \{1, 2, ..., k\}$

Over a discretized object, the multiple integral can be rewritten:

$$\sum_{X \in object} X_1^{n_1} X_2^{n_2} ... X_k^{n_k} dm$$, where $dm$ is the unitary mass element in the discrete space.

In a more compact way, it will be written: $M_{(object)} \left( \prod_{i=1,k} X_i^{n_i} \right)$

Then the discrete mass of an object will be: $M_{(object)}(1)$

From the moments of order 1, it can be deduced the object gravity center by:

$$XG_i = M_{(object)}(X_i) / M_{(object)}(1), \ i \in \{1, 2, ..., k\}$$

It will be further neglected to refer to the object as the support domain of moments.

The centered values of moments of order 2 will be obtained according to the following manner:

$$\text{if } x_i = X_i - XG_i, \; i \in \{1, 2,, k\} \text{ then}$$
$$M(x_i x_j) = M(X_i X_j) - XG_i M(X_j) - XG_j M(X_i) + XG_i XG_j M(1)$$

The centered moments of order 2 enable to compute the rotation matrix which is defining the axes of the object Eigen reference frame. This one is built from the inertia matrix of the object:

$$In_{k \times k}(i, j) = M(x_i x_j), \quad i \in \{1, 2, ..., k\}, j \in \{1, 2, ..., k\}$$

It is a squared definite positive matrix that can be rewritten after diagonalization:

$$In_{k \times k} = V^T \Lambda V$$

where $\Lambda_{k \times k}(i, j) = M(u_i, u_j) \;/\; M(u_i^2) \geq 0$ et $M(u_i u_j) = 0$ for $i \neq j$

and $\forall i \in \{1, 2, ..., k-1\}, \quad M(u_i^2) \geq M(u_{i+1}^2)$

is the matrix of the inertia axes of the object represented in its Eigen reference frame, and where $V$ is the matrix of the object Eigen vectors ($V^T V = VV^T = I$) enabling to move from the centered reference frame to the Eigen reference frame of the object by rotation.

At this stage, an object is assimilated to its inertia ellipsoid. The Eigen vectors of its reference frame are then defined more or less π.

For removing the uncertainty about the direction of inertia axes, it should be used the moments of order 3.

The centered values of moments of order 3 can be written:

$$M(x_i x_j x_m) = M(X_i X_j X_m) - XG_i M(X_j X_m)$$
$$- XG_j M(X_i X_m) - XG_m M(X_i X_j) + XG_i XG_j M(X_m)$$
$$+ XG_i XG_m M(X_j) + XG_j XG_m M(X_i) - XG_i XG_j XG_m M(1)$$

The expression of the values of moments of order 3 restricted to the Eigen directions is then:

$$M(u_i^3) = \sum_j \sum_m \sum_n v_{ji} v_{mi} v_{ni} M(x_j x_m x_n)$$

where $v_{ji}, v_{mi}, v_{ni}$ are the components of the Eigen vector $v_i$ of the matrix $V$ of reference frame change.

The moments of order 3 restricted to the Eigen directions $M(u_i^3)$ are the asymmetries of the object according each of its Eigen axes. They are measuring the object eccentricity along each axis.

The uncertainty of the axis directions is removed by orientating the axes in the direction of the strongest eccentricity, that is to say in such a meaning where:

$$M(u_i^3) \geq 0, \quad \forall i \in \{1, 2, ..., k\}$$

That is by replacing $v_i$ by its opposite $-v_i$ in the matrix $V$ of reference frame change, when $M(u_i^3) < 0$.

In such a way, it has been got the matrix of translation, rotation for a displacement of the object tree in its Eigen reference frame described in homogenous coordinates:

$$\begin{bmatrix} U \\ 1 \end{bmatrix} = \begin{bmatrix} V & -XG^T \\ 0 & 1 \end{bmatrix} \begin{bmatrix} X \\ 1 \end{bmatrix}$$

For obtaining a representation invariant to any scaling, it is only needed to normalize the coordinates with respect to the main inertia axis:

$$\begin{bmatrix} U \\ 1 \end{bmatrix} = \begin{bmatrix} \dfrac{1}{M(u_1^2)} V & -XG^T \\ 0 & 1 \end{bmatrix} \begin{bmatrix} X \\ 1 \end{bmatrix}$$

This representation might take in account different scaling factors according each axis, but it would not get any reality concerning rigid body mechanics.

Under $L^1$, all the black nodes observed at any precision in a tree are compact and close sub-sets of $[0,1[^k$.

The binary tree of a set of $[0,1[^k$ is then a countable compact covering of this set. It is finite when it is examined at a finite precision. So any measure calculated on the binary tree of a set will be equal to the sum of measures calculated over each compact sub-set of this covering.

Therefore, it will be possible to compute hierarchically the moments linked to each grey or black nodes of the tree and to sum them up on the tree traversal return at every non terminal node the contribution of each terminal node in order to deduce the measures of the object modeled by the tree.

It will be now described a recursive method for the calculus of generalized moments on the blocks visited during the traversal of a tree representing an object on $[0,1[^k$.

To the tree nodes are associated parallelepipeds with edges parallel to the axes:

$$[x_1', x_1''[\times [x_2', x_2''[\times \cdots \times [x_k', x_k''[$$

That will be assimilated under $L^1$ with:

$$[x_1', x_1'']\times [x_2', x_2'']\times \cdots \times [x_k', x_k'']$$

And that will be written as:

$$\prod_{i=1,k}[x_i', x_i'']$$

These blocks have got for generalized moments:

$$M_{\left(\prod_{i=1,k}[x_i',x_i'']\right)}\left(\prod_{i=1,k}X_i^{n_i}\right) = \prod_{i=1,k}\frac{1}{n_i+1}\left(x_i^{''n_i+1} - x_i^{'n_i+1}\right)$$

In a binary tree at precision $p$, a terminal node will get for moment:

$$M_{\left(\prod_{i=1,k}[x_{ig},x_{id}]\right)}\left(\prod_{i=1,k}X_i^{n_i}\right) = \prod_{i=1,k}\frac{1}{n_i+1}\left(x_{id}^{n_i+1} - x_{ig}^{n_i+1}\right)$$

If the support domain $\prod_{i=1,k}[x_{ig}, x_{id}]$ is divided according to $x_j$ into sub-sets of same size, it will be got the following values:

$$M_{\left(x_{jg},\frac{x_{jg}+x_{jd}}{2}\right)\prod_{\substack{i=1,k\\i\neq j}}[x_{ig},x_{id}]}\left(\prod_{i=1,k}x_i^{n_i}\right) = \frac{1}{n_j+1}\left(\left(\frac{x_{jg}+x_{jd}}{2}\right)^{n_j+1} - x_{jg}^{n_j+1}\right)\prod_{\substack{i=1,k\\i\neq j}}\frac{1}{n_i+1}\left(x_{id}^{n_i+1} - x_{ig}^{n_i+1}\right)$$

Knowing that:

$$\left(\frac{x_{jg}+x_{jd}}{2}\right)^{n_j+1} - x_{jg}^{n_j+1} = \frac{1}{2^{n_j+1}}\sum_{m=0}^{n_j+1}C_{n_j+1}^m x_{jg}^m x_{jd}^{n_j+1-m} - x_{jg}^{n_j+1}$$

and that:

$$(1+1)^{n_j+1} = \sum_{m=0}^{n_j+1}C_{n_j+1}^m = 2^{n_j+1}$$

It can be written:

$$M_{\left(\left[x_{jg},\frac{x_{jg}+x_{jd}}{2}\right]\prod_{\substack{i=1,k\\i\neq j}}[x_{ig},x_{id}]\right)}\left(\prod_{i=1,k}x_i^{n_i}\right) = \sum_{m=0}^{n_j}\frac{C_{n_j+1}^m}{2^{n_j+1}} \times \frac{n_{j+1}-1}{n_{j+1}} \times x_{jg}^m M_{\left(\prod_{i=1,k}[x_{ig},x_{id}]\right)}\left(x_j^{n_j-k}\prod_{i=1,k}x_i^{n_i}\right)$$

Likewise:

$$M_{\left(\left[\frac{x_{jg}+x_{jd}}{2},x_{jd}\right]\prod_{\substack{i=1,k\\i\neq j}}[x_{ig},x_{id}]\right)}\left(\prod_{i=1,k}x_i^{n_i}\right) = \sum_{m=0}^{n_j}\frac{C_{n_j+1}^m}{2^{n_j+1}} \times \frac{n_{j+1}-1}{n_{j+1}} \times x_{jd}^m M_{\left(\prod_{i=1,k}[x_{ig},x_{id}]\right)}\left(x_j^{n_j-k}\prod_{i=1,k}x_i^{n_i}\right)$$

For lightening the presentation, it will thereafter only be taken in account of the dividing interval for displaying the support application domains.

The development of these expressions concerning the generalized moments gives:

At order 0:

$$M_{\left(\left[x_{jg},\frac{x_{jg}+x_{jd}}{2}\right]\right)}(1) = \frac{1}{2}M_{\left([x_{jg},x_{jd}]\right)}(1)$$

$$M_{\left(\left[\frac{x_{jg}+x_{jd}}{2},x_{jd}\right]\right)}(1) = \frac{1}{2}M_{\left([x_{jg},x_{jd}]\right)}(1)$$

At order 1:

$$M_{\left(\left[x_{jg},\frac{x_{jg}+x_{jd}}{2}\right]\right)}(X_i) = \frac{1}{2}M_{\left([x_{jg},x_{jd}]\right)}(X_i)$$

$$M_{\left(\left[\frac{x_{jg}+x_{jd}}{2},x_{jd}\right]\right)}(X_i) = \frac{1}{2}M_{\left([x_{jg},x_{jd}]\right)}(X_i)$$

$$M_{\left(\left[x_{jg},\frac{x_{jg}+x_{jd}}{2}\right]\right)}(X_j) = \frac{1}{4}M_{\left([x_{jg},x_{jd}]\right)}(X_j) + \frac{1}{4}x_{jg}M_{\left([x_{jg},x_{jd}]\right)}(1)$$

$$M_{\left(\left[\frac{x_{jg}+x_{jd}}{2},x_{jd}\right]\right)}(X_j) = \frac{1}{4}M_{\left([x_{jg},x_{jd}]\right)}(X_j) + \frac{1}{4}x_{jd}M_{\left([x_{jg},x_{jd}]\right)}(1)$$

And so on until the order 3.

After having determined the moments of each black $2^k$-ant of the tree describing the object, the object moments are then equal to:

$$M_{(object)}\left(\prod_{i=1,k} X_i^{n_i}\right) = \sum_{black\ 2^k-ant \in object}\left(\prod_{i=1,k} X_i^{n_i}\right)$$

Computed up to the order 3, the generalized moments provide after centering and normalization:

- at the order 0, the hypervolume of the component,
- at the order 1, its gravity center,
- at the order 2, its inertia axes,
- at the order 3, the signature of its reference frame and its asymmetries according to every axes.

So the calculus of generalized moments allows to settle a component building invariant to linear positive transformations that might apply on the component in affine geometry, its Eigen tree.

## VI.2 – Pattern recognition

After the centering and the normalization of the moment list of a component, this one is linked to a vector of measures independent of similarities and symmetries, its attributes:

$$\left(M(u_1^2),\cdots,M(u_k^2),M(u_1^3),\cdots,M(u_k^3)\right),$$

where:

$$M(u_1^2) = 1 \text{ or } M(u_i^2) > 0 \text{ et } M(u_i^3) > 0$$

Therefore the component could be represented by a vector of $2k-1$ measures.

For implementing a procedure of pattern recognition relying on this representation, it must be built a learning set made from $n$ experiences:

$$\{(f_i, v_i)\}, i = 1, n,$$

where $v_i$ is the attribute vector of the $i$-th experience and $f_i$ the label attributed to the experience by a teacher coming from a finite set of possible interpretations (supervised learning).

A procedure of hierarchical classification enables then to classify according to the Hausdorff distance each new observation expressed using this new form of representation, over the partition of the interpretations.

In the case of $2^k$-trees, the learning step consists in building the chromatic tree of dimension 2k-1 modeling the set $\{(f_i, v_i)\}$.

During the recognition step, every new observation will be interpreted (classified), by finding back the label to which she is linked in the learning tree.

Finding back this label consists, when the learning tree has been modeled by a chromatic tree, in building the $2^{2k-1}$-tree associated to the attribute vector, then in performing the Boolean intersection of the tree with the learning pyramid and in reading the label $f_i$ extracting from the resulting tree.

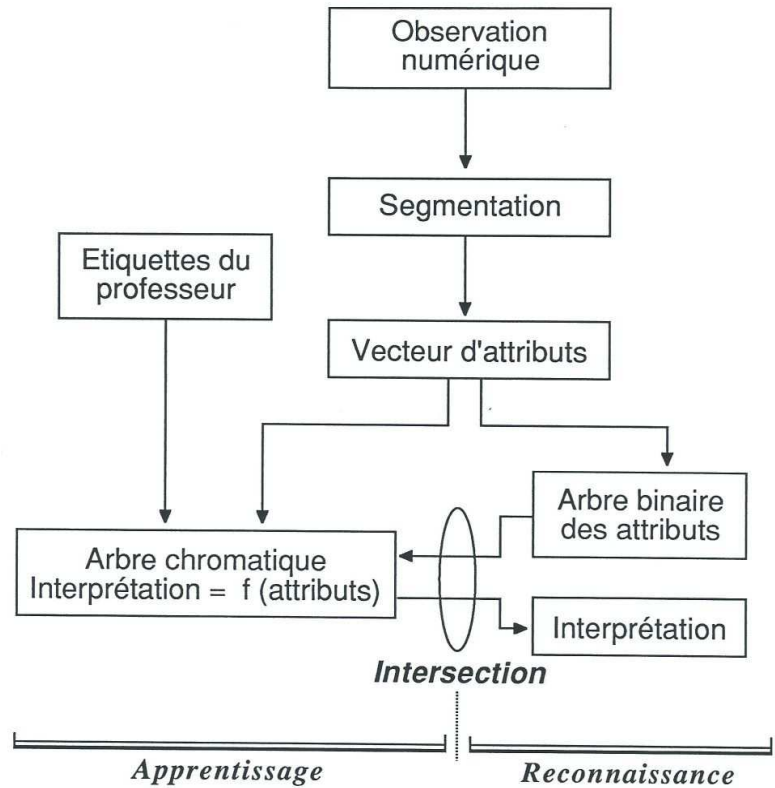

**Figure 17 : Spectral pattern recognition**

It has been seen that the moment calculus of a connected component enables to rebuild the component in its Eigen reference frame. It is then possible to propose a procedure finer than the spectral method that has just been presented:

- by building the component Eigen tree with the help of the information provided by the moment calculus
- by replacing the interpretation base relying on attributes, by the union of the Eigen components labeled with the same interpretations (cf. fig. 18 et 19).

It has been viewed that a correlation measure between trees is standing with the mass (hypervolume) of the exclusive-or of two trees. If a new experience is now appearing:

- by computing its Eigen tree,
- by performing the exclusive-or of this tree with the data base,
- by measuring the remaining mass of the base for each label of the learning set,

- the label whose mass is the lightest is then the label the best correlated with the new experience.

It can be noticed that the learning base, being the Boolean union of the Eigen trees of the learning set, is no more bulky than an Eigen component.

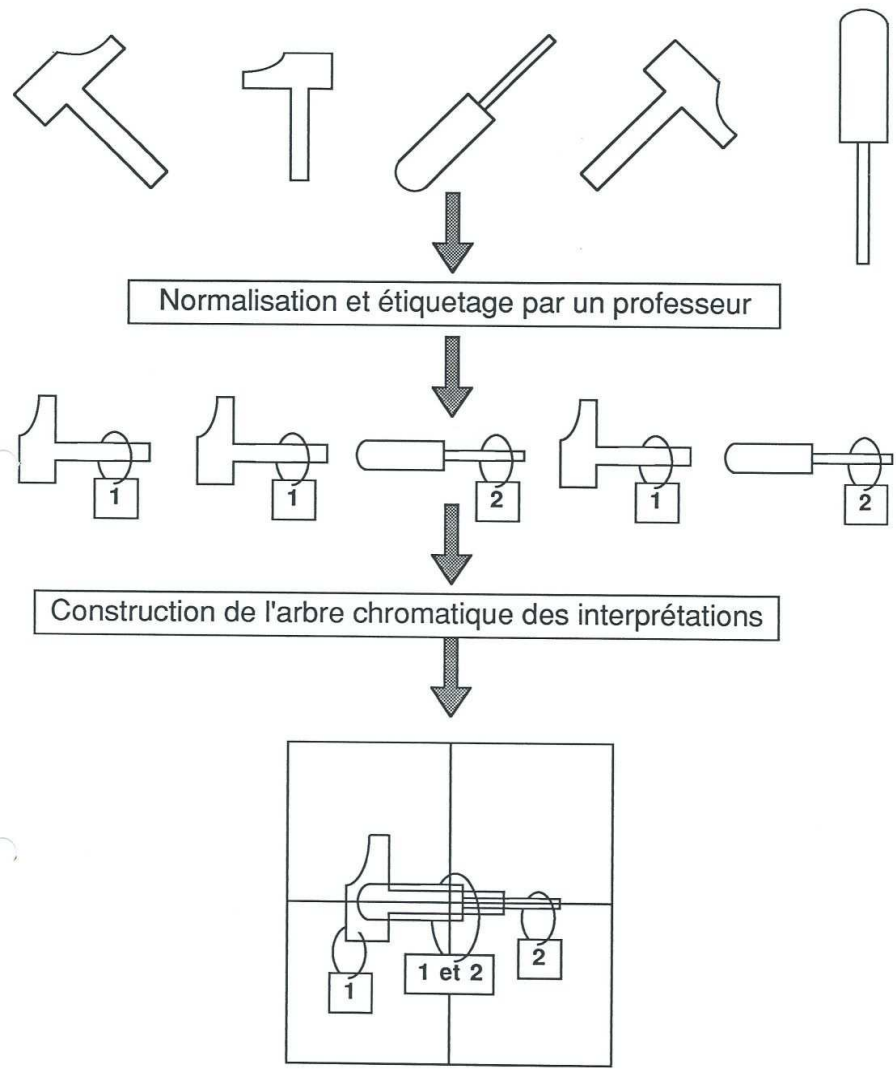

**Figure 18 : Data base of the Eigen trees**

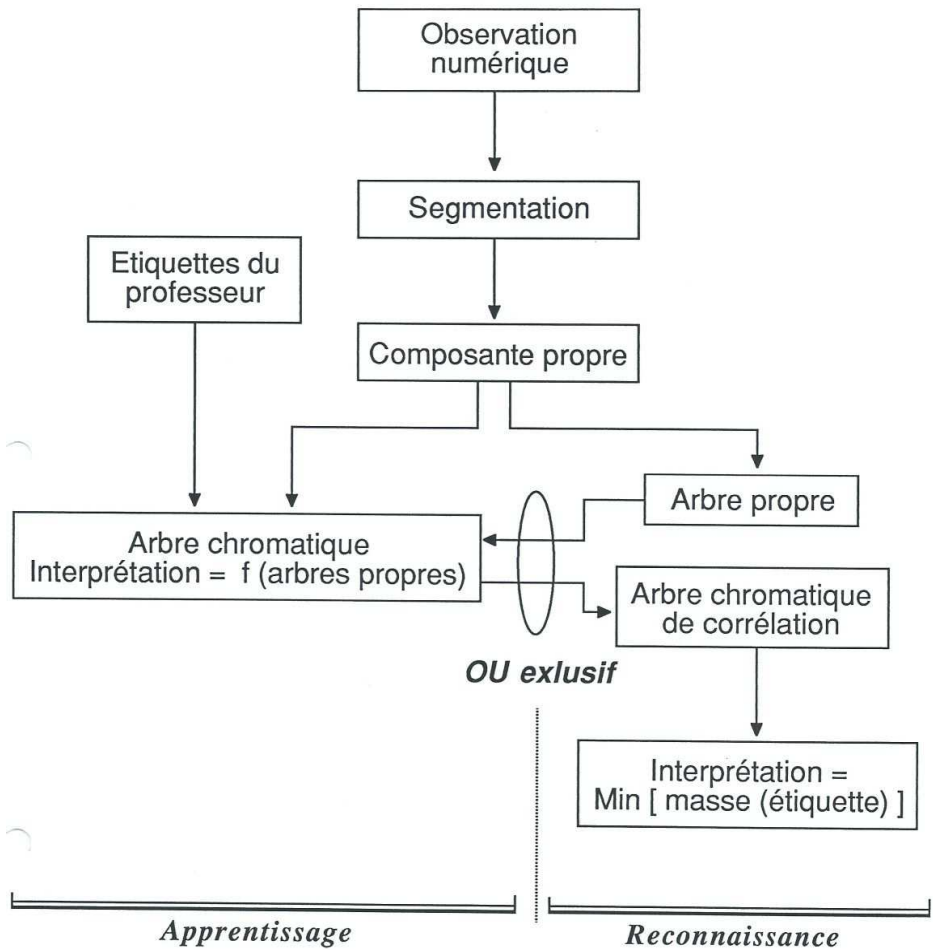

**Figure 19 : Correlative pattern recognition**

# Conclusion

This presentation about $2^k$-trees modeled by binary trees has shown that the results obtained by the past on quaternary and octernary trees can be extended to spaces of any dimension.

The proposed tree building method is compatible with disorderly and overcrowded data streams.

Its inductive limit extension enables to avoid any previous normalization of the data, while insuring the comparability of trees generated independently the ones from the others.

The computation of the homographic transform of a $2^k$-tree has clarified the properties of convex analysis needed for performing such a transformation, used by some authors in spaces of two or three dimensions.

As it has been widely pointed out, the calculus of attributes based on the generalized moments enables to:

- localize in position and in orientation, an object in its observation space ;
- deduce a vector of measures invariant to similarities ;
- propose for each object a description independent from its observation space, its Eigen tree.

It has been described the principles of segmentation relying on two distances $d_1$ et $d_\infty$ induced by the structure of mesh associated to $2^k$-trees.

These principles enable to define the notion of object as a connected component. According to this approach, they enable to implement perception procedures of objects without any recovering in affine geometry.

Two methods of unsupervised pattern recognition have been described.

A third unsupervised one can be designed in introducing a supplementary segmentation stage applied in the attribute space in the case of the spectral pattern recognition: the membership class is the connected component reached after the learning phase.

These techniques mix several different approaches commonly met in statistical data analysis:

- Bayesian, where the tree of a connected component modeled a probability distribution ;
- by partitioning while proposing two segmentation methods ;
- by hierarchical classification knowing that the representation model is driven by an ultrametric distance, the Hausdorff distance;
- factorial analysis, by the means of generalized moments.

In the framework of the mid-term study which has enabled to perform these works, some other aspects of the question have been also studied:

- the computation of the convex hull of a multidimensional object (checking the linear separation of classes) ;
- the insertion, the extraction of manifolds parallel to the reference frame (parallel slices) ;
- the homotopic transformations (erosion, dilation, median filtering) ;
- the calculus of the integral equation system of a continuous manifold by approximation with variable order polynomials and the building of the atlas of maps of a piecewise continuous manifold;
- the search of Hamiltonian paths on data sets (Peano-Hilbert scanning) ;
- the recognition of patterns partially hidden in projective geometry by building dual trees ;
- the analysis of parallel architecture computers agreeing with this kind of algorithmic (mainly parallel architectures based on multiple interconnection networks).

# Glossary

**adjacency :** direct link between two points of a graph, proximity relation in discrete geometry, the most often deducted from the distance used on the space of interest (cf. distance, neighborhood, connectivity).

**aggregation :** method in statistical analysis that enables to gather data sets or to compute measures over these sets (aggregating indicators used for instance in the building of dashboards).

**Boolean algebra :** set algebra, all the operations that can be the result of a combination of elementary set operations (assertion, negative, union, intersection, exclusion, difference).

**Borel algebra :** set made from all the parts of a space that are enclosed for the algebraic operations that can be applied on – if the number of applied operations is countable then it is a sigma-algebra. The sigma-algebras are at the foundation of the theories of measure and probabilities– in the present case, they are semi-open intervals coming from the regular decomposition of the unitary space down to a finite integer precision for a Borel algebra and a sigma-algebra when the precision is approaching the infinite.

**relational algebra :** all the operations that can apply on a data set and which made from the combination of a finite number of elementary relational operations (union, intersection, jointure, Cartesian product,...).

**algorithm :** computing procedure applying the one after the other a finite number of simple computing rules enabling to perform a complex function – it can be distinguished sequential algorithms from parallel ones where the order followed for processing operations and data is not the same.

**altimetry :** technique for measuring altitudes, by extension the set of altitudes regularly sampled on a planar cartographic support.

**analysis :** decomposition of a whole into its parts (cf. synthesis)

**factor analysis :** statistical data analysis technique based on the search and the visualization of data according to their first factors (main inertial axes of the data cloud).

**image analysis :** all the image decomposition techniques into elementary subsets, that are functional values regularly sampled on planar supports.

**linear analysis :** mathematical analysis technique focusing on the decomposition of functions at first differential order in multidimensional spaces and on their handling using matrix algebra, enabling for instance to directly solve inverse systems.

**spatial analysis :** part of statistical analysis which is dealing with random systems evolving within space.

**statistical analysis :** analysis of random systems containing a large number of elements, in which it is tried to distinguish exogenous factors from endogenous ones with the help of several observations on a same event.

**structural analysis :** part of systems analysis dealing with the decomposition of systems into subsystems and the study of exchanges that they may have between them using exchanges matrices or interconnection graphs.

**time-series analysis :** part of statistical analysis which is dealing with random systems evolving with time.

**approximation :** numerical method enabling to reproduce a data set with a given error - usually its upper bound is controlled, and it provides a mathematical model which is fitting to an initial data set.

**archiving :** gather, classify and preserve pieces of information.

**first level archiving :** archiving of new pieces of information in a raw format without applying any processing

**second level archiving :** archiving of old or of compressed pieces of information.

**attitude :** orientation information about an object in its frame of reference : in the plane, angle of the main inertia axis with the abscissa axis of the reference frame; in the space, matrix of object Euler angles in its reference frame.

**attributes :** measures computed on the whole set of data defining an object (they are also called characteristics) ; they enable to apply statistical pattern recognition techniques for identifying objects.

**attributes in differential geometry:** measures applying on surface data (in other words functions evolving according to a planar support), as the surface mean, the tangent plane, the curvature tensor.

**attributes in discrete geometry :** measures applying on volume data (in other words functions evolving according to a tridimensional support), as the generalized moments, the inclosing rectangle, the form factor.

**attributes in statistical analysis :** measures applying on object populations from which can be inferred a mean vector, a variance-covariance matrix, etc...

**authentication :** pattern recognition process where it is not tried to identify a shape, but only to confirm that it is the right one that it is expected to find (the plausibility of a proposed label for an observed shape is only checked).

**Eigen axes :** or also axes of inertia, they enable to assimilate an object with its inertia ellipsoid and provide information about the attitude of an object around its gravity center; they are deduced from its generalized moments and enable to define the Eigen frame of reference of an object.

**median axes :** set of points that are at equal distance from the borders of an object ; it is a topological description that can vary according to the used distance, also named the skeleton of an object.

**principal axes :** they are the axes that are gathering the main part of inertia among all the Eigen axes, in statistical data analysis, they enable to reduce the observing space down to a sub-space of lower dimension and to observe data with losing only a small part of the information (descriptive analysis).

**boundary :** set of points belonging to a connected component that are both connected to the inside and the outside of the component.

**cataloging :** identifying, indexing and summarizing information.

**characteristics :** attributes.

**classes** : sub-sets sharing the same property.

**classification :** defining classes.

**computer architecture classification :** Flynn has settled a computer architecture classification relying on the degree of multiplicity of instruction and data streams (SISD, MISD, SIMD MIMD), to which can be linked a programming model named SPMD that can apply on the two different situations of massive parallelism (SIMD, SPMD over MIMD architecture).

**hierarchical classification :** classification methods leading to the building of nested partitions, usually modeled by tree-like structures where parts can be compared between them using an ultrametric distance.

**partitioning-based classification :** classification methods based on the building of a partition over a data set.

**supervised classification :** to set up classes where the way how to classify is shown by a teacher, this one labels a learning set with the labels of the expected classes.

**thematic classification :** from multispectral image analysis in remote sensing, classification of luminescence vectors of images points, so as to attempt and to recover the physical nature of elements covering the ground (snow, water, fields, forests, urban areas,...).

**unsupervised classification:** to set up classes where the way how to classify is straightly computed using a classification method without the help of a teacher for labeling a learning set.

**compactness :** for a shape, it is a state which is similar to some neighborhoods or to a small union of neighborhoods belonging to the topology of the descriptive space – in planar image analysis, the compactness of an object can be measured with the help of form factor which is the ratio of the object surface on the square of its perimeter : the closer to one it is, the more compact the form is.

**connected component :** subset made from connected points (equivalence class according to a relation of connectivity or adjacency).

**homogeneous component :** subset of data of same kind.

**compression :** action enabling to reduce the size of an information set (analysis) ; the original set is retrieved by decompression (synthesis).

**lossless compression :** compression method for which it is sure to retrieve the whole original information, it is usually similar to perform a format conversion.

**lossy compression :** compression method for which it is not sure that at decompression time original information can be retrieved without any loss.

**asynchronous computer :** parallel architecture computer where each processor has at its disposal a control unit applying on its own instruction stream, enabling to perform different programs over the data set to be processed.

**synchronous computer :** parallel architecture computer where all the processors share a same control unit applying on the instruction stream, implying the performance of a single program simultaneously on all over the data set to be processed.

**high-performance computing :** performance of algorithms where numerical computing is prevailing (first instance, solving high-dimensional differential equation systems).

**parallel computing :** performance of algorithms on specific architecture computers, favoring parallel processing on similar computing units enabling to decrease their processing time.

**vector computing :** performance of algorithms on specific architecture computers, favoring the parallel processing of given arithmetic operations for data appearing under the form of numerical vectors.

**connectivity :** link or straight relation between two points in a set ; usually proximity relation in a metric space, neighborhood relation in a discrete space.

**contour :** boundary of an object in the plane ; it has the property to be visited only once and using only one way (border following).

**format conversion :** operation enabling to convert a data set from one data format to another one.

**thematic conversion :** during or after a thematic classification, operation enabling to convert a multispectral image into a thematic image whose pseudo-colors depict the theme linked to the luminescence vectors identified during the classification step.

**connectivity degree :** number of elements to which an element can be related.

**expansion degree :** rank up to which a function can be approximated using a limited expansion (degree 0, constant function - degree 1, linear - degree 2, quadric - degree 3, cubic).

**singularity degree** : expansion degree for which a function cannot be approximated in a given location of its support – its functional or its differential tensor up to this degree is not locally continuous.

**intrinsic dimension :** roughly, dimension of the space in which an object can be mapped using a continuous transform in such a way that it would recover a main part of this space; so the intrinsic dimension of a point is zero, the value for a curve is one, and two for a surface – using piecewise linear fitting, these elements correspond to points, lines and surfaces in cartography.

**distance :** measure enabling to value the proximity of two points (metric) or two point sets (ultrametric), print a particular topology over the corresponding space and rely on a specific mesh, when this space is discretized.

**metric distance :** usually on the $d_k$ distances(k-th root of the sum of the p-th power of the absolute value of the coordinates difference), three from them are the most useful, $d_1$ and $d_\infty$ that can be discretized on regularly sampled meshes according to each dimension of the space and $d_2$ that can be discretized in the plane using a hexagonal mesh.

**ultrametric distance :** a main focus is applied on Haussdorf or Hamming distances, which are compliant with regular meshes built using $d_1$ and $d_\infty$, and generate over these spaces rougher topologies, but for which a partition can be hierarchically handled (for instance by using trees of degree $2^k$).

**division :** resolution principle used by the algorithms that are following the "divide and conquer" paradigm which consists in decomposing a problem into sub-problems when they cannot be straightly solved; it prints out a hierarchical structure over the processed data for solving the corresponding problem (a quick sort, a fast Fourier transform or the building of a convex hull are following this method in order to provide algorithms optimal in computing time).

**epigraph :** half-space above a curve or a surface.

**expansion :** operation enabling to extend a discrete data set so as to produce a continuous covering of the support space – this operation includes an hypothesis concerning the data distribution – used at degree 0 for building a continuous covering of a labeled space during a statistical learning.

**convolutional filtering :** numerical filtering of a signal based on the use of a convolution operator, which degree enables to perform or an integration operation, either a derivative operation on the numerical data.

**topological filtering :** or morphological filtering, based on the recursive use of two basic transform, erosion and dilatation – it applies on binary data (support space) or on multi-valued data (functional space).

**forest :** collection of un-connected graphs, the more often trees representing connected components of the space.

**polygon filling :** format conversion enabling to transform a vectorized representation model into a cellular representation model.

**garbage-collector :** process enabling to reorganize the free space of a variable-legnth memory allocation system.

**analytic geometry :** specific field of geometry which is focused on the handling of geometric shapes using analytic expansions, that is to decompose forms into primitives (approximation using expansions in series for representing curves or surfaces).

**computational geometry** : specific field of geometry which is dealing with algorithms enabling to extract and to transform geometrical structures registered in numerical data sets.

**constructive geometry :** synthesis of geometrical objects by using algebraic combinations (with the help of Boolean operations completed with some affine transformations) of complex geometric shapes.

**differential geometry :** extension of analytic geometry where the coefficients implied in series expansion are interpreted as local differential operators and with whose help it is possible to extract attributes for locally characterizing the analyzed shape.

**discrete geometry :** all the properties and the transformations induced by the sampling of a geometrical shape sampled over a regular mesh or not – they are mainly neighborhood properties and topological transformations (connected components, erosion-dilation, inside-border-outside).

**projective geometry :** all the techniques that allow numerical data to be projected into a sub-space; two specific case should be particularly noticed, the one in perspective geometry where it is tried to know what can be observed from a given viewpoint of the space and the other one dealing with information reduction in statistical data analysis where it is only tried to project data from the initial observation space into the principal sub-space associated to the gravity center and to the inertial properties of its data cloud – in the first case, a homography is used, in the second one it is an orthography (viewpoint placed at the infinite).

**homotopy :** continuous transformation in topology.

**hypograph :** half-space under a curve or a surface.

**distribution hypothesis :** assumed form of the statistical distribution of numerical data when certain statistical data analysis techniques are used (for instance the factor analysis techniques data should obey to the Gaussian distribution hypothesis) – the distribution hypothesis is replaced by a hypothesis of topological structuring of the observed space for the methods relying on topological analysis.

**identification :** fit the coefficients of a parametric model to a numerical data set, or classify an observation according to an already known classification.

**generalized image :** superposition of different models of data representation in order to describe a digital image (cellular representation, segmentation, decomposition into geometrical structures, relational organization of homogeneous components).

**monochrome image :** image digitized according a single luminous frequency.

**multichrome image :** image digitized according several luminous frequencies, where each radiometric contribution is kept in a differentiated manner so as to form a collection of digital images.

**panchromatic image :** image digitized according several luminous frequencies, but where only the union of these responses are kept into a single digital image.

**pseudo-color image :** cellular image where the value of a picture element does not represent a luminescence but the label of a connected component to which belongs the luminescence or a vector of luminescence for a multispectral image.

**hashing-based index :** way of managing information based on the direct algebraic transformation of the binary representation of this same information – it is quck manner for retrieving pieces of information, by enabling both distribution holes and reference collisions in the storage space.

**linear index :** way of managing information based on the information sort and its sequential access - for retrieving a piece of information, it is nearly necessary to visit all them all, at the opposite this addressing system does not create any hole or collision.

**hierarchical index :** way of managing information based on the information sort and the hierarchical access to it - for retrieving a piece of information, only a branch of the indexing tree must be visited – if this system does not produce any collision, reaching a minimum time access needs to dynamically reorganize the addressing index in order to keep a balanced tree

**dynamic index :** tree-like index, computed every time when the information is updated so as to obtain a minimal time access to the information – the reorganization time of the index is increasing with the size of the information to be retrieved and may become unaffordable in massive data management.

**static index :** tree-like index, statically defined at the generation of the data set and left un-modified when the information is updated – access time is under-optimal, but no computing time is required for a data reorganization, this kind of indexing better fits to massive data management.

**indexing :** method for storing some data into a data set.

**cartographic information :** planar representation or result of a planar projection applied on a geographic information.

**geographic information :** information which is about the Earth description.

**geometric information** : information about the description of objects in the spaces where they are observed or represented.

**interpretation :** word used in the restricted meaning of classifying, that is to link a label to an observation (descriptive meaning, but not explaining).

**invariance** : invariance to geometric transforms, property that should verify measures applied on objects in order to enable the use of statistical analysis techniques so as to classify them, if it is not the case the localization information must systematically linked to the object to be registered – it is difficult to check it in projective spaces and it can only be certified when observation points are nearly at the infinite (enough distance is needed for globally recognize a shape).

**labeling :** action which consists in putting labels (marking) on experiments and which is performed after the partitioning of a set in order to classify data.

**learning :** step in pattern recognition where it is tried to tune a process in pattern recognition, usually by showing what to do on a data set chosen in advance.

**localization :** act enabling to locate in position and in attitude an object in its observation space.

**data base management** : all the means enabling to record, to update, and to access according to various query schemes to a set of digital data.

**data file management :** all the means enabling to record, to update, and to access according to various query schemes to a set of digital data for which the organization is explicit.

**data file management :** all the means enabling to record, to update, and to access according to various query schemes to a set of digital data lying on a specific physical medium as a computer disk.

**massive data management :** all the means enabling to record digital data set whose size is very higher than those that are commonly handled.

**geometric match :** describes the fact that it is possible to move from one approximant to its neighbor one on their common boundary, continuously up to their approximation degree (or nearly) – this property may usefully apply when the modeled surface is regular.

**matching** : in structural pattern recognition, procedure enabling to find an isomorphism (a strong correspondence) between two sub-graphs of two graphs.

**Lebesgue measure:** measure that matches with the volumes of the Cartesian products of bounded intervals from $R$ into $R^n$. This measure is at the foundation of the calculation of Lebesgue integral, which is defined as the limit value of two series made from the products of two layered functions bordering a piece-wise continuous function and converging at the infinite towards a single finite value.

**similarity measure :** comparison measure between two pieces of information inducing a pre-ordering over a collection of numerical data, technique used in statistical analysis enabling to sort the answers according to a plausibility order after a query by example applied over a numerical data base.

**global memory :** process enabling to address in a single step and to access at the whole or a part of the memory belonging to a multiprocessor system.

**memory addressing :** in the framework of a partitioned memory system, shared or distributed, it provides global addressing by using contiguous, interlaced addressing or any other strategy mixing these two approaches (especially by shuffling address bits).

**memory granule :** smallest unit that can be allocated in a mapped memory or a virtual memory – when the length of this one is fixed, it is theoretically not necessary to use a garbage-collector for re-organizing the free memory (that is not fully right for data handled on a disk).

**mapped memory** : memory divided into fixed length units, named pages, enabling to implement a memory allocation mechanism without any re-organization and a fast addressing system working by address translation.

**segmented memory :** memory that can be divided into variable-length segments, enabling to implement a memory allocation mechanism needing a periodic re-organization of its free space.

**virtual memory :** process enabling to address and to access a memory volume which size is bigger than the one physically accessible by a processor – it uses a secondary memory with which it communicates by granules exchange.

**mesh :** topological structuring made from the discretization of a numerical data set over a support space – when the discretization is applied at a fixed step over all the support axes, a regular mesh is produced from which partial storing orders or visiting paths can be extracted from the data (for instance, a grid) – at the contrary, only a proximity information can be kept (for instance, a Delaunay triangulation).

**parallel programming model :** methods used for programming on parallel computers – two from them are more quoted : data-driven programming on SIMD architecture and message-passing programming on MIMD architecture.

**generalized moments :** integral measures performed on objects enabling to provide mutually independent measures, as pieces of information about their localization in the observation space and values invariant with similarities – computed by integrating monomials on the object supports, developed up to a given degree.

**multidimensional :** the fact to be described in a space of several dimensions – two in image analysis, three in the real space, several in statistical data analysis or in operations research.

Multidimensional Hierarchical Modeling : Tome 1                                                         Page 83

**multiresolution :** the fact to describe a signal as a continuous series of representations observed at different resolutions evolving according to a given geometrical law.

**multiscale :** the fact to be described at different scales of representation (macro-, meso-, micro-scale) – enables to distinguish the interactions according to their ranges in structural analysis.

**multispectral :** the fact to be decomposed, for a signal, over a basis of elementary signals, usually characterized by an own frequency, whose re-composition enables to render the initial signal

**navigation :** ability to move oneself inside a numerical data base regarding a given frame of reference

**neighborhood :** set of points adjacent to a given point, its shape varies according the topology used for describing the space.

**interconnection network :** data exchange system in a parallel architecture computer, allowing all the processors (distributed system) or all the processors and all the memories (shared system) to simultaneously communicate with respect to some constraints linked to the topology of the interconnection network (regular accesses without any addressing conflict).

**object :** set of homogeneous data about which can be performed an interpretative process

**observation :** information capture about the presence of objects inside a given space(with the help of a perception system).

**algebraic operation :** sequence of several elementary operations that can be expressed using a formal language.

**Boolean operation :** operation applying on sets and providing a new set.

**inductive limit operation :** operation from which it can be deduced the reference frame of the expected result knowing those of its operands when it is applied to bounded spaces.

**geometric operation** : geometric transform of one set into another one – troubles may be encountered when it implies a re-sampling of data belonging to discrete sets.

**topologic operation :** continuous transform of the shape of an object, enabling to preserve the connectivity degree of its graph.

**optimization :** numerical method enabling to localize the optimum (minimum or maximum) of a function – used in approximation in order to reduce the error between the data and the numerical models used to fit them.

**constrained optimization :** special case in optimization, where the search is applied on a restrictive domain of the support representing the constraints to be respected – used in some given fields of operations research, especially for solving problems of finite resources allocation.

**order** : way to arrange elements comparatively between each other – rank of an element in a list of elements.

**orientation :** attitude of an object around a position in a given space, measured relatively to the axes of a reference frame.

**parallelization :** act making executable an algorithm whatever it may be sequential or parallel on a parallel computer - act to be distinguished from the design of parallel algorithms: in the first case, it is a translating task, in the second one it is a design task.

**partitioning :** division of a set into several sub-sets.

**data partitioning :** distribution of a data set over several memory units.

**active perception :** information capture about an environment by applying some action on it - usually by emitting a signal and by analyzing the signal sent back due to the interaction of the former signal with the environment to be observed.

**passive perception :** information capture about an environment without applying any action on it – the analysis is restricted to natural signals produced by the environment to be observed.

**perspective :** deformation generated on a space when it is observed from a given point.

**piecewise :** decomposition of a function support so as to enable its fitting with serial developments of low-degree functions.

**planimetry :** planar projection of the whole set of geometrical structures lying on a given terrain.

**planisphere :** result of the projection of a sphere on a plane – in practice, map representing the two terrestrial or celestial hemispheres.

**control points :** support points on which are lying the approximants belonging to a piecewise regular approximation of a surface - usually the support vertices of the approximant in the used mesh.

**fitting polynomial :** polynomial approximating a data collection over a mesh element – the fitting is usually applying without any error with the data collection.

**position :** location where an object is standing in a space.

**precision :** detail measure, measure of the smallest element taken in account – it can so be distinguished the capture precision of observations, from the calculation precision for the operators that can be applied on these observations, from the representation or modeling precision which corresponds to the precision used to keep in memory these observations.

**variable precision :** property of operations that can be applied at any precision, that are able to process numerical objects roughly as well as in a detailed manner – this property influences the processing time of these operations and enables to develop des operators working in controlled time.

**primitive :** set of functions or simple shapes on which a geometrical object can decomposed – when it is dealing with the boundary of an object, it is a function set that enables to approximate the boundary with a piecewise description – usually line segments for a planar shape (linear approximation).

**projection :** operation enabling to transfer a data set from a space of a given dimension to a space of a lower dimension

**proximity :** order relation induced by a distance over a discrete data set.

**pseudo-color / theme :** label connected to a class of radiometric vectors in a multispectral image – when is connected an explanation of physical nature to the class, then it is a theme.

**query :** search for an information expressed by using a question, a request.

**query by address :** request made by telling where is put the awaited information.

**query by example :** request made by showing what kind of information is sought.

**content-based query :** close to the query by example, in the meaning that is tried to rebuild the pieces of information that should be in the data management system.

**structured query :** request written using an algebraic language describing the relations that should share the simple elements of the answer.

**approximate reasoning :** perform plausibility calculations about the happening of given events by managing distinctly the facts that can occur with those that cannot occur.

**geometric reasoning :** solve geometrically geometrically referenced problems.

**operations research :** mathematical field connected to statistical data analysis which is focused on solving problems rather than describing them, by especially relying on constrained optimization and on graph theory.

**recognition :** step in pattern recognition where a recognition procedure is enabled to retrieve the classification or the labeling that has been stored after a learning step – two main approaches can be distinguished : <u>the statistical pattern recognition</u> which is a global approach based on attributes computing and on statistical classification, and the <u>structural pattern matching</u> which a local approach based on primitives computing and on graph matching.

**geometric rectification :** corrective action of the geometry of a remote sensing image in order to counterbalance satellite movements along its trajectory relatively to the observed scene.

**radiometric rectification :** sensors correction according to known parameters with regards to their individual operating behavior; it is usually performed an inter-sensor equalization.

**recording :** register a piece of information for keeping it in such a way enabling to further render it.

**reduction :** operation enabling to reduce the size of a data set, without losing or with losing part of the information – the statistic data analysis tries by decreasing the dimension of the descriptive space or by identifying equivalence classes (quotient spaces) to implement techniques of problem reduction.

**reference frame :** geometric frame enabling to localize itself in a space with its bounds along each dimension when the space is bounded – these data are used to define all the geometric transforms that enable to move and to turn itself in this space.

**capture reference frame :** reference frame linked to the space in which signals coming from the observation space are received (the plane for images).

**Eigen reference frame :** reference frame linked to the analyzed object.

**observation reference frame :** reference frame linked to the space where are lying the observer and the observed object (tridimensional, for the world in which we are moving and looking at).

**principal reference frame :** reference frame linked to the first inertia axes of a numerical data set.

**support reference frame :** reference frame linked to the support space of a functional : reference frame linked to the signals which are digitalized (mono- / multi-chromatic signals for an image).

**universal reference frame :** reference frame enabling to merge several partial observations performed in a same observation space from different points of view.

**region :** connected points set having the same intrinsic dimension as its description space – inside and boundary of a connected component.

**regular :** in approximation theory, characterizes a curve or a surface that can be modeled by an expansion in limited series – the continuity order of an original object is not always known, at the opposite the order of its singularities or irregularities can be retrieved by looking at its responses when differential operators are applied on it - otherwise, due to their regularity, a curve or a surface can be approximated by piecewise polynomials of low degree.

**regularization :** approximation method used in presence of locally singular data ; the problem is usually solved by using a variational frame where smoothing constraints are introduced in the error to minimize.

**hidden parts removal :** operation enabling to remove all the parts of an observed object that cannot be viewed by an observer located at a given point: this operation applies when is displayed on a flat screen a tridimensional scene viewed in perspective, or when the visibility graph related to an observation system placed on a digital terrain model.

**data representation :** a same data set can be represented using different manners – for turning from one representation format into another one, it is necessary to perform a format conversion procedure – according to the processing to be applied on the data set, its duration varies with the concerned representation formats.

**boundary-based representation :** representation model where a data set is described by its boundary (for instance, the contour of an object or the vectorization of a shape).

**region-based representation :** representation model where a data set is described for all the information which is inside or on its boundary (for instance, all the cellular representations or the trees of degree $2^k$).

**surface-based representation :** representation model where a data set is described by surfaces from which is only kept an expression in the form of a piecewise polynomial approximation (triangulated surface).

**volume-based representation :** representation model where a data set is directly described in its representation space (matrices of volume elements, trees of degree $2^k$).

**cellular representation :** representation model where the data are values sampled over a regular mesh and are handled as tables or matrices (images, altimetries).

**vectorized representation :** representation model where a data set is described by curves from which is only kept an expression in the form of a piecewise linear approximation (vectors or line segments).

**sampling-resampling :** discretization of a set of continuous data, usually performed by using a fixed-length step on each dimension of the function and its support during their capture; meshed data may be highly altered by a geometric transform – vectorized data better cope with these transforms than cellular data.

**scale :** the evolution of a complex system can be observed at different resolution scales: global, median and local (macro-, meso-, micro-scale) – the high- and mid-scales are usually got by data aggregation – the interactions between subsets are observed according to their range at the corresponding scale: for instance statistical pattern recognition is only focused only global phenomena applying on object shapes, at the opposite structural pattern recognition is only focused on the local configuration of primitives coming from the decomposition of the same shapes.

**magnitude scale :** the size of data sets handled in numerical data archiving can be huge – each scale needs its own data recording, updating and querying functions – by cascading them, it is then possible to act in an interoperable manner between the different levels of data description – these scales are currently related between them by factors of thousands in physical systems.

**segment :** connected component produced by a segmentation.

**segmentation :** search of connected components satisfying to a given predicate (for instance, the isocoloration predicate in image analysis).

**shape :** object structure.

**simple shape :** simple geometric structure – in the plane, the contour of such an object can be represented by a piecewise linear approximation.

**complex shape :** collection of simple geometric structures sharing proximity relations.

**signaling :** description intended to make known the existence of an information set.

**similitude :** geometric transform composed of translations, rotations and scaling.

**singular :** irregular, which is not regular.

**sort :** arrange a series of elements according to an order relation.

**bounded space :** all the data sets are presumed to be observed in spaces with finites bounds, usually known (frame of reference) or computed (operations in inductive limits).

**functional space :** evolution domain of the observed data – for images, the luminescence or the vector of radiometric response.

**support space :** space in which is spread data, which when it is discretized induces a mesh over this space.

**complex structure :** arrangement of simple structures.

**geometric structure** : arrangement of geometric primitives.

**relational structure :** arrangement of structures, composed using a relational algebra (notably sharing some given position relations).

**simple structures :** arrangement of parts into a set.

**summarization :** to reduce the information size while only keeping main information.

**synthesis :** re-composition of a whole from its parts (operation inverse to the analysis).

**distributed memory system :** parallel architecture system where the memories are locally linked to the processors and where the communications are only established between the processors.

**partitioned memory system :** parallel architecture system that can be decomposed into autonomous sub-systems enabling several users or applications to use it simultaneously without any interference.

**shared memory system :** parallel architecture system where the memories are globally accessible to all the processors and where the communications can be established between processors and memories.

**neighborhood system :** way for nesting neighborhoods in each point of a space and inducing a new topology over this space (for instance, multiresolution operations and hierarchical classification that induce an ultrametric topology).

**thesaurus :** normalized information directory enabling to perform a classification (in the case of image cataloging, it is built from the classification of geometrical shapes after a thematic conversion).

**thickness :** smallest diameter of an object – when an object is analyzed with a precision of the same range than its thickness, its intrinsic dimension may become lower (object of unitary thickness).

**thinning :** topological operation enabling to erode a set while preserving the connectivity order of its points – the infinite iteration provides its median set.

**topology :** study of geometrical properties that can be preserved using continuous deformations.

**address translation :** field-based addressing system enabling to quickly localize a memory granule in a bank memory and/or a multistage memory system (main - auxiliary), and to re-compute a physical address from a virtual address.

**tree :** computer data structure and directed graph of a unitary input degree in graph theory.

**complete tree :** tree whose all branches are fully developed (for instance, pyramidal trees).

**balanced tree :** tree developed in such a way that all its branches are nearly all developed down to the same depth and consequently providing the fastest accesses to the data of a given set.

**binary tree :** tree of degree 2, enabling to model a finite data set belonging to a bounded space of dimension 1.

**quadtree :** tree of degree 4, enabling to model by regular decomposition a finite data set belonging to a bounded space of dimension 2.

**octtree : :** tree of degree 8, enabling to model by regular decomposition a finite data set belonging to a bounded space of dimension 3.

**tree of degree** $2^k$ **:** tree enabling to model by regular decomposition a finite data set belonging to a bounded space of dimension k.

**pyramidal tree:** tree of degree $2^k$, where all the branches are developed down to its maximum precision for modeling data.

**vectorization :** piecewise linear approximation of a linear structure or the contour of a surface structure.

# Annex : hierarchical modeling algorithms

## 1. Data structure management

addstr :        address of the structure to be processed

succ :          successor in a list

lftson :        left son of a node in a tree

rgtson :        right son of a node in a tree

lftlgt :        length of left sub-tree

rgtlgt :        length of right sub-tree

## 1.1. Deletion of any data structure

PROCEDURE kddest(addstr)

BEGIN

    /* analysis of its type and call of the associated deletion operator */

    IF (*tree type*(addstr)) THEN DO

        CALL kddebt(addstr)

        *delete root*(addstr)

    END

    ELSE IF ((*linear list type*(addstr)) OR (*circular list type*(addstr)))

    THEN CALL kddeli(addstr)

END

## 1.2. Deletion of a structure of list type

PROCEDURE kddeli(addstr)

BEGIN

    IF (*circular list type*(addstr)) THEN convert the circular list into a linear list

    /* list traversal and deletion of sub-structures */

    succ <- *link*(addstr)

    WHILE (NOT *nil*(succ)) DO

        IF (NOT *simple variable* (succ)) THEN CALL kddest(value(succ))

        succ <- *link*(succ)

    END

    /*effective deletion of the list*/

    IF (NOT *nil*(succ)) THEN *delete list*(addstr)

END

## 1.3. Deletion of a structure of tree type

PROCEDURE kddebt(addstr)

BEGIN

    IF (NOT *terminal*(addstr)) THEN DO

        /* depth-first tree traversal */

        CALL kddebt (*left son*(addstr))

        CALL kddebt (*right son*(addstr))

        /* merge on return path */

        iso-coloring of left and right sons

        *merge*(addstr)

    END

    IF (valued tree (addstr)) THEN DO

        /* deletion of the structure linked to the node */

        CALL kddest(*value*(addstr))

        *type* (addstr) <- unvalued tree

    END

END

## 1.4. Copy of any data structure

FUNCTION kdcpst(addstr)

BEGIN

    /* analysis of its type and call of the associated copy operator */

    IF (*tree type*(addstr)) THEN kddpst <- kdcpbt(addstr)

    ELSE IF ((*linear list type* (addstr)) OR (*circular list type* (addstr)))

    THEN kdcpst <- kdcpli(addstr)

END

## 1.5. Copy of a structure of list type

FUNCTION kdcpli(addstr)

BEGIN

    IF (*circular list type* (addstr))

    THEN convert the circular list into a linear list

    /* list traversal and duplication */

    IF (NOT *nil*(addstr)) THEN kdcpli <- *queue head* (*type*(addstr))

    succ <- *link*(addstr)

    WHILE (NOT *nil*(succ)) DO

        IF (*simple variable* (succ)) THEN *insert in queue* (kdcpli, *value* (succ), *type* (succ))

        ELSE *insert in queue* (kdcpli, kdcpst (*value* (succ)), *type* (succ))

        succ <- *link*(succ)

    END

    IF (*circular list type* (addstr))

    THEN convert into circular list the linear lists addstr et kdcpli

END

## 1.6. Copy of a structure of tree type

```
FUNCTION kdcpbt(addstr)
BEGIN
    IF (terminal(addstr)) THEN DO
        /* copy of a terminal node */
        kdcpab <- tree (link(addstr), type(addstr))
    END
    ELSE DO
        /* depth-first tree traversal */
        lftson <- kdcpbt (left son (addstr))
        rgtson <- kdcpbt (right son (addstr))
        kdcpbt <- sub-trees union (lftson, rgtson)
    END
    IF (unvalued tree (adrstr)) THEN value (kdcpab) <- value (adrstr)
    ELSE DO
        /* copy of the structure linked to the node */
        value(kdcpbt) <- kdcpst (value(addstr))
        type(kdcpbt) <- type(addstr)
    END
END
```

## 1.7. Length of any data structure

FUNCTION kdlgst(addstr)

BEGIN

    /* analysis of its type and call of the associated calculus operator */

    IF (*tree type*(addstr)) THEN kdigst <- kdlgbt(addstr)

    ELSE IF ((*linear list type* (addstr)) OR (*circular list type* (addstr)))

    THEN  kdlgst <- kdlgli(addstr)

END

## 1.8. Length of a structure of list type

FUNCTION kdlgli(addstr)

BEGIN

    IF (*circular list type* (addstr))

    THEN convert the circular list into a linear list

    IF (NOT *nil*(addstr)) THEN kdlgli <- 1 ELSE kdlgli <- 0

    succ <- *link*(addstr)

    WHILE (NOT *nil*(succ)) DO

      IF (*simple variable* (succ))

      THEN kdlgli <- kdlgli + 1

      ELSE kdlgli <- kdlgli + kdlgst(*value*(succ)) + 1

      succ <- *link*(succ)

    END

    IF (*circular list type* (addstr))

    THEN convert into a circular list the list addstr

END

## 1.9. Length of a structure of tree type

FUNCTION kdlgbt(addstr)

BEGIN

    IF (*terrninal* (addstr)) THEN kdlgbt <- 1

    ELSE DO

      /* depth-first tree traversal */

      lftlgt <- kdlgbt (*left son* (addstr))

      rgtlgt <- kdlgbt (*right son* (addstr))

      kdlgbt <- lftlgt + rgtlgt +1

    END

    IF (*valued tree* (addstr)) THEN DO

      /* length calculus of the structure linked to the node */

      $\text{kdlgbt} <- \text{kdlgbt} + \text{kdlgst}(value(\text{addstr}))$

    END

END

## 2. Tree generation by vector addition

root :          root of the tree to be enriched

vecthd :        head of the integer vector to be added

vector:         head of the real vector to be added

dimens :        dimension of the modeling space

precis :        computation precision

depth :         computation depth

level :         level reached in the tree

side:           descent side in the tree

minroo :        minimum coordinates list of the root

maxroo :        maximum coordinates list of the root

xmin :          minimum coordinate of the root

xmax :          maximum coordinate of the root

xvec :          vector coordinate to be added

xctr :          center coordinate of the root

## 2.1. Addition of an integer vector to a tree

PROCEDURE kdaivt(root ,vecthd, dimens, precis)

BEGIN

    /* initialization of computing parameters */

    depth <- dimens * precis, level <- 0

    /* recursive computing unit */

    PROCEDURE kdaivt (root, vecthd , level)

    BEGIN

        IF (level=depth) THEN *blacken* (root)

        ELSE_DO

            /* tree descent driven by the vector coordinates */

            side <- extraction of most significant bit of current coordinate,

                right binary shift of this one,

                and coordinate circular shift of vector(vecthd)

            IF (*terminal*(root)) THEN *fission*(root)

            CALL kdaivt(*son*(root, side), vecthd, level + 1)

            *merge* (root)

        END

    END

END

## 2.2. Addition of a normalized real vector to a tree

PROCEDURE kdarvt(root, vector, dimens ,precis)

BEGIN

    /* initialization of computing parameters */

    depth <- dimens * precis, level <- 0

    {minroo , maxroo}<- heads of vectors({(0., 0., ..., 0.), (1., 1., ..., 1.)})

    /* recursive computing unit */

    PROCEDURE kdarvt (root , minroo, maxroo, vector, level)

    BEGIN

        IF ((level <> depth) AND (NOT *black*( racine)) THEN DO

            /* tree descent driven by the vector coordinates */

            {xmin, xmax, xvec} <- coordinates extraction from heads{minroo,maxroo,vector }

            xctr <- (xmin + xmax)/2.

            IF (xvec<xctr)

            THEN xmax <- xctr,  side <- <u>left</u>

            ELSE xmin <- xctr,  side <- <u>right</u>

            *vector queue insert* ({minrac, maxrac, vector }, {xmin, xmax, xvec})

            IF (*terminal*(root)) THEN *fission*(root)

            CALL kdarvt (*son* (root, side), minroo, maxroo, vector, level + 1)

        END

        ELSE *blacken* (racine)

        IF (NOT *terminal* (racine)) THEN *merge*(racine)

    END

    *delete vectors*({minroo , maxroo})

E N D

# 3. Boolean operations

| | |
|---|---|
| root : | root of the operand tree |
| root1 : | root of the first operand tree |
| root2: | root of the second operand tree |
| dimens : | dimension of the modeling space |
| precis : | computation precision |
| depth : | computation depth |
| level : | level reached in the tree |
| lftson : | left son of the root of the resulting tree |
| rgtson : | right son of the root of the resulting tree |

## *3.1. Assertion of a binary tree*

FUNCTION kdass(racine, dimens, precis)

BEGIN

    /* initialization of computing parameters */

    depth <- dimens * precis, level <- 0

    /* recursive computing unit */

    FUNCTION kdass(root, level)

    BEGIN

       IF ((NOT *terminal* (root)) AND (level <> depth)) THEN DO

          /* depth-first tree traversal */

          lftson <- kdass(*left son*(root), level + 1)

          rgtson <- kdass(*right son*(root), level + 1)

          kdass <- *sub-trees union*(lftson, rgtson)

       END

       ELSE DO

          /* assertion of the node reached in the tree */

          IF (*white* (root))

          THEN kdass <- *tree* (<u>white</u>)

          ELSE kdass <- *tree* (<u>black</u>)

       END

       /* merge of sons' nodes on return path */

       IF (NOT *terminal* (kdass)) THEN *merge* (kdass)

    END

END

## *3.2. Negation of a binary tree*

FUNCTION kdnot(root, dimens, precis)

BEGIN

    /* initialization of computing parameters */

    depth <- dimens * precis, level <- 0

    /* recursive computing unit */

    FUNCTION kdnot(root, level)

    BEGIN

      IF ((NOT *terminal* (root)) AND (level <> depth)) THEN DO

        /* depth-first tree traversal */

        lftson <- kdnot(*left son* (root), level + 1)

        rgtson <- kdnot(*right son* (root) ,level + 1)

        kdnot <- *sub-trees union* (lftson, rgtson)

      END

      ELSE DO

        /* negation of the node reached in the tree */

        IF (NOT *white*(racine))

        THEN kdnot <- *tree* (<u>white</u>)

        ELSE kdnot <- *tree* (<u>black</u>)

      END

      /* merge of sons' nodes on return path */

      IF (NOT *terminal* (kdnot)) THEN *merge*(kdnot)

    END

END

## 3.3. Union of two binary trees

FUNCTION kdunio(root1, root2, dimens, precis)

BEGIN

    /* initialization of computing parameters */

    depth <- dimens * precis, level <- 0

    /* recursive computing unit */

    FUNCTION kdunio (root1, root2, level)

    BEGIN

        IF ((NOT *terminal* (root1)) OR (NOT *terminal* (root2))) AND (level <> depth)) THEN DO

            /* parallel depth-first traversal of the two trees*/

            lftson <- kdunio (*left son* (root1), *left son* (root2), level + 1)

            rgtson <- kdunio (*right son* (root1), *right son* (root2), level + 1)

            kdunio <- *sub-trees union* (lftson, rgtson)

        END

        ELSE DO

            /* union of the reached nodes */

            IF ((*white* (root1)) AND (*white* (root2)))

            THEN kdunio <- *tree* (<u>white</u>)

            ELSE kdunio <- *tree* (<u>black</u>)

        END

        /* merge of sons' nodes on return path */

        IF (NOT *terminal* (kdunio)) THEN *merge* (kdunio)

    END

END

## 3.4. Intersection of two binary trees

```
FUNCTION kdintr(root1, root2, dimens, precis)
BEGIN
    /* initialization of computing parameters */
    depth <- dimens * precis, level <- 0
    /* recursive computing unit */
    FUNCTION kdintr(root1, root2, level)
    BEGIN
        IF ((NOT terminal (root1)) OR (NOT terminal (root2))) AND (level <> depth)) THEN DO
            /* parallel depth-first traversal of the two trees */
            lftson <- kdintr(left son (root1), left son (root2), level + 1)
            rgtson <- kdintr(right son (root1), right son (root2), level + 1)
            kdintr <- sub-trees union (lftson, rgtson)
        END
        ELSE DO
            /* intersection of the reached nodes */
            IF ((NOT white(root1)) AND (NOT white(root2)))
            THEN kdintr <- tree (black)
            ELSE kdintr <- tree (white)
        END
        /* merge of sons' nodes on return path */
        IF (NOT terminal (kdintr)) THEN merge(kdintr)
    END
END
```

## 3.5. Exclusion of two binary trees

```
FUNCTION kdexcl(root1, root2, dimens, precis)
BEGIN
    /* initialization of computing parameters */
    depth <- dimens * precis, level <- 0
    /* recursive computing unit */
    FUNCTION kdexcl(root1, root2, level)
    BEGIN
        IF ((NOT terminal (root1)) OR (NOT terminal (root2))) AND (level <> depth)) THEN
        DO
            /* parallel depth-first traversal of the two trees */
            lftson <- kdexcl(left son (root1), left son (root2), level + 1)
            rgtson <- kdexcl(right son (root1), right son (root2), level + 1)
            kdexcl <- sub-trees union (lftson, rgtson)
        END
        ELSE DO
            /* exclusion of the reached nodes */
            IF ((white (root1)) EX OR (white (root2)))
            THEN kdexcl <- tree (black)
            ELSE kdexcl <- tree (white)
        END
        /* merge of sons' nodes on return path */
        IF (NOT terminal (kdexcl)) THEN merge (kdexcl)
    END
END
```

## 3.6. Difference of two binary trees

FUNCTION kddiff(root1, root2, dimens, precis)

BEGIN

    /* initialization of computing parameters */

    depth <- dimens * precis, level <- 0

    /* recursive computing unit */

    FUNCTION kddiff(root1, root2, level)

    BEGIN

        IF ((NOT *terminal* (root1)) OR (NOT *terminal* (root2))) AND (level <> depth)) THEN DO

            /* parallel depth-first traversal of the two trees */

            lftson <- kddiff(*left son* (root1), *left son* (root2), level + 1)

            rgtson <- kddiff(*right son* (root1), *right son* (root2), level + 1)

            kddiff <- *sub-trees union* (lftson, rgtson)

        END

        ELSE DO

            /* difference of the reached nodes */

            IF ((NOT *white* (root1)) AND (*white* (root2)))

            THEN kddiff <- *tree* (<u>black</u>)

            ELSE kddiff <- *tree* (<u>white</u>)

        END

        /* merge of sons' nodes on return path */

        IF (NOT *terminal* (kddiff)) THEN *merge* (kddiff)

    END

END

# 4. Handling of slices parallel to the axes

| | |
|---|---|
| spacro : | root of the space to be processed |
| dimesp : | dimension of the modeled space |
| slicro : | root of the slice to be inserted |
| dimslc : | dimension of the slice space |
| slvcro : | tree root of the slice coordinates vector |
| codslc : | dimension of the tree of the slice coordinates (codimension of the slice in the space to be processed) |
| slcvec : | vector head of the slice axes (for each coordinate of the space to be processed, it indicates if the axis is a slice axis or not) |
| precis : | computation precision |
| slaxis : | axis in the process of analysis belonging to the slice axes vector |
| depth : | computation depth |
| level : | level reached in the tree |
| lftson : | left son in the slice |
| rgtson : | right son in the slice |
| intrsc : | intersection of the slice with the slice coordinates |

## 4.1. Extraction of a slice parallel to the axes

FUNCTION kdexsl(spacro, dimesp, slvcro, dimslc, slcvec, precis)

BEGIN

    /* initialization of computing parameters */

    depth <- dimesp * precis, level <- 0

    slaxis <- *link*(slcvec)

    /* recursive computing unit */

    FUNCTION kdexsl (spacro, slvcro, slaxis, level)

    BEGIN

        IF ((NOT *white* (slvcro)) AND (level <> depth)) THEN DO

            /* parallel depth-first traversal of the two trees */

            IF (*value* (slaxis))

            THEN lftson <- kdexsl (*left son* (spacro), *left son* (slvcro), *link* (slaxis), level + 1)

            ELSE lftson <- kdexsl (*left son* (spacro), slvcro, *link* (slaxis), level + 1)

            IF (*value* (slaxis))

            THEN rgtson <- kdexsl (*right son* (spacro), *right son* (slvcro), *link* (slaxis), level+1)

            ELSE rgtson <- kdexsl (*right son* (spacro), slvcro, *link* (slaxis), level + 1)

            kdexsl <- *sub-trees union* (lftson, rgtson)

    END

    ELSE DO

        /* slice extraction: */

        /* intersection of the space with the slice coordinates vector*/

        IF ((NOT *white* (racesp)) AND (NOT *white* (racdcp)))

        THEN kdexsl <- *tree* (black)

        ELSE kdexsl <- *tree* (white)

    END

    /* merge of sons' nodes on return path */

    IF (NOT *terminal* (kdexsl)) THEN *merge* (kdexsl)

    IF (*value* (slaxis)) THEN delete in the slice tree kdexsl the coordinates along the slice axis

E N D

E N D

## 4.2. *Insertion of a slice parallel to the axes*

PROCEDURE kdinsl(spacro, slicro, dimslc, slvcro, codslc, slcvec, precis)

BEGIN

    /* initialization of computing parameters */

    depth <- (dimslc + codslc) * precis,  level <- 0

    slaxis <- *link* (slcvec)

    /* recursive computing unit */

    PROCEDURE kdinsl (spacro, slicro, slvcro, slaxis, level)

    BEGIN

      IF ((NOT *white*( slicro)) AND (level <> depth)) THEN DO

        /* parallel depth-first traversal of the three trees */

        IF (*value* (slaxis))

        THEN CALL kdinsl (*left son* ((spacro), slicro, *left son* (slvcro), *link* (slcvec), level + 1)

        ELSE CALL kdinsl (*left son* (spacro), *left son* (slicro), slvcro, *link* (slcvec), level + 1)

        IF (*value* (slaxis))

        THEN CALL kdinsl (*right son* (spacro), slicro, *right son* (slvcro), *link* (slcvec), level + 1)

        ELSE CALL kdinsl (*right son* (spacro), *right son* (slicro), slvcro, *link* (slcvec), level + 1)

    E N D

```
    ELSE DO
        IF (level=depth) THEN DO
            IF (NOT terminal (spacro)) THEN put at terminal state the sub-tree spacro
            /* slice insertion: */
            /* computation of the intersection of the slice with the slice coordinates vector */
            IF ((NOT white (slicro)) AND (NOT white (slvcro)))
            THEN intrsc <- black
            ELSE intrsc <- white
            /* union of the space with the result of the intersection */
            IF ((white (spacro)) AND (white (intrsc)))
             THEN whiten (spacro)
            ELSE blacken (spacro)
        END
    END
    /* merge of sons' nodes on return path */
    IF (NOT terminal (spacro)) THEN merge (spacro)
  END
END
```

## 5. Building of the tree of a polyhedron

polyed :         list of the polyhedron vertices

minhyp :         list of the polyhedron lower faces

maxhyp :         list of the polyhedron higher faces

dimens :         dimension of the modeled space

precis :         computation precision

depth :          computation depth

level :          level reached in the tree

{polspc, minspc, maxspc} :     vertices and faces of a block coming from the regular dividing de of the modeled space

inters :         intersection indicator of two polyhedrons

incp12 :         inclusion of the first polyhedron into the second one

incp21 :         inclusion of the second polyhedron into the first one

{polye1, minhp1, maxhp1} :     vertices and faces of the first polyhedron

{polye2, minhp2, maxhp2} :     vertices and faces of the second polyhedron

nudim :          number of the processed dimension

nulspc :         inclusion indicator of a polyhedron into a hyper-plane

negspc :         inclusion indicator of a polyhedron into a negative half-space

posspc :         inclusion indicator of a polyhedron into a positive half-space

hyplan :         hyper-plane partitioning the spce onto two half-spaces

nuvert :         vertex number

order :          butterfly size

nubut :          bufferfly number

dimhom :         dimension of the space in homogenous coordinates

nuplan :         plane number in the faces list

{pollft, minlft, maxlft} :   vertices and faces of the left half-polyhedron

{polrgt, minrgt, maxrgt} :   vertices and faces of the right half-polyhedron

## 5.1. Building of the tree of a polyhedron defined by its vertices and its faces

FUNCTION kdpolt(polyed, minhyp, maxhyp, dimens, precis)

BEGIN

/* building initialization */

/* computation of the vertices and the faces of the block linked to the root */

polspc <- unitary space vertices (dimens)

minspc <- unitary space lower hyperplanes (dimens)

maxspc <- unitary space higher hyperplanes (dimens)

depth <- dimens * precis, level <- 0

/* recursive computing unit */

FUNCTION kdpolt ([polyed, minhyp, maxhyp,] polspc, minspc, maxspc, level)

BEGIN

/* hierarchical traversal of the inside and the boundary of the polyhedron */

intersection evaluation of the block linked to the node and of the polyhedron (polspc, minspc, maxspc, polyed, minhyp, maxhyp)

IF ((level <> depth) AND ((intersection) AND (block $\not\subset$ polyhedron))) THEN DO

/* the block linked to the node is on the boundary of the polyhedron */

{pollft, polrgt} <- vertex-based polyhedron division (polspc)

{minlft , maxlft, minrgt, maxrgt}<- face-based polyhedron division (minspc, maxspc)

lftson<- kdpolt (pollft, minlft, maxlft , level + 1)

rgtson <- kdpolt (polrgt, minrgt, maxrgt, level + 1)

kdpolt <- *sub-trees union* (lftson, rgtson)

*merge* (kdcvap)

END

ELSE_DO

                /*covering at the required precision or inclusion of the block linked to the node into the polyhedron*/

                IF (((level = depth) AND (intersection)) OR (block $\subset$ polyhedron))

                THEN kdpolt <- *tree* (<u>black</u>).

                ELSE kdpolt <- *tree* (<u>white</u>)

            END

            delete block {polspc, minspc, maxspc}

        END

END

## 5.2.   Intersection evaluation of two convex polyhedrons

PROCEDURE kdiecp(inters, incp12, incp21, polye1, minhp1, maxhp1, polye2, minhp2, maxhp2, dimens)

BEGIN

    /* the intersection is not validated when all the vertices of a polyhedron belong to the half-space related to one of the hyperplanes belonging to the other polyhedron */

    /* a polyhedron is included into the other polyhedron, when all its vertices belong to the internal half-spaces defined by the hyperplanes of the other polyhedron */

    inters <- <u>true</u>

    /* examination of the faces of the first polyhedron */

    Incp12 <- <u>true</u>

    /* examination of the lower hyperplanes of the first polyhedron */

    FOR nudim = 1 TO dimens DO

        position evaluation of a polyhedron compared to a hyperplane (polye2, minhp1(nudim), inside hyperplane, left space, right space)

        /* the left space is outside the poylhedron, the right space is inside*/

        IF (NOT inside hyperplane) THEN_DO

            IF (NOT right space) THEN incp12 <- <u>false</u>

            IF (left space) THEN inters <- <u>false</u>

        END

END

/* examination of the higher hyperplanes of the first polyhedron */

FOR nudim = 1 TO dimens DO

    position evaluation of a polyhedron compared to a hyperplane (polye2, maxhp1(nudim), inside hyperplane, left space, right space)

    /* the left space is inside the polyhedron, the right space is outside */

    IF (NOT inside hyperplane) THEN_DO

        IF (NOT left space) THEN incp12 <- false

        IF (right space) THEN inters <- false

    END

END

/* examination of the faces of the second polyhedron */

incp21 <- true

/* examination of the lower hyperplanes of the second polyhedron */

FOR nudim = 1 TO dimens DO

    position evaluation of a polyhedron compared to a hyperplane (polye1, minhp2(nudim), inside hyperplane, left space, right space)

    IF (NOT inside hyperplane) THEN_DO

        IF (NOT right space) THEN incp21 <- false

        IF (left space) THEN inters <- false

    END

END

/* examination of the higher hyperplanes of the second polyhedron */

FOR nudim = 1 TO dimens DO

    position evaluation of a polyhedron compared to a hyperplane (polye1, maxhp2(nudim), inside hyperplane, left space, right space)

    IF (NOT inside hyperplane) THEN_DO

        IF (NOT left space) THEN alors incp21 <- false

        IF (right space) THEN alors inters <- false

    END

END

overriding of inclusion indicators when not any intersection

END

## 5.3. *Position evaluation of a polyhedron compared to a hyperplane*

```
PROCEDURE kdpohp(nulspc, negspc, posspc, polyed, hyplan, dimens)
BEGIN
    /*test initialization */
    nulspc, negspc, posspc <- true
    /* examination of polyhedron vertices */
    FOR nuvert = 1 TO 2**dimens DO
        posit <- 0.
        /* examination of vertex coordinates */
        FOR nudim = 1 TO dimens DO
            posit <- posit + (polyed(nuvert, nudim) * hyplan(nudim))
        END
        posit <- posit + hyplan(nudim + 1)
        IF (posit=0.) THEN nulspc <- false
        IF (posit>0.) THEN negspc <- false
        IF (posit<0.) THEN posspc <- false
    END
END
```

## 5.4. Vertex-based polyhedron division into two half-polyhedrons

```
PROCEDURE kddivp(polyed, pollft, poldrt, order, dimens)
BEGIN
    /* update of the size of decomposition butterfly lattices */
    nudim <- dimens – LOG2(order)
    nudim <- MOD(nudim + 1, dimens) + 1
    order <- 2**(dimens - nudim)
    /* polyhedron analysis according butterflies of order vertices */
    FOR nubut = 1 TO MOD(2**dimens, order) DO
        /* division of the butterfly vertices */
        FOR nuvert = ((nubut - 1) * order) TO (nubut * order) DO
            nuver1 <- nuvert
            nuver2 <- nuvert + order
            /* coordinates division */
            FOR nudim = 1 TO dimens DO
                pollft(nuver1, nudim) <- polyed (nuver1, nudim)
                polrgt(nuver2, nudim)
                    <- (polyed (nuver1, nudim) + polyed (nuver2, nudim))/2.
                polrgt (nusom1, nudim) <- pollft (nuver2, nudim)
                polrgt (nusom1, nudim) <- polyed (nuver2, nudim)
            END
        END
    END
END
```

## 5.5. Face-based polyhedron division into two half-polyhedrons

PROCEDURE kddivh(minhyp, maxhyp, minlft, maxlft, minrgt, maxrgt, npldiv, dimens)

BEGIN

    /*space dimension in homogenous coordinates*/

    dimhom <- dimens + 1

    /* polyhedron division in two halves separated by the median hyperplane of the npldiv-th lower and higher faces */

    FOR nuplan = 1 TO dimhom DO

        /* division of polyhedron faces */

        FOR nudim = 1 TO dimhom DO

            /* division of faces into homogenous coordinates */

            minlft(nuplan, nudim) <- minhyp(nuplan, nudim)

            IF (nuplan=npldiv) THEN DO

                maxlft(nuplan, nudim) <- (minhyp(nuplan, nudim)+maxhyp(nuplan, nudim))/2.

                minrgt(nuplan, nudim) <- maxlft(nuplan, nudim)

            END

            ELSE DO

                maxlft(nuplan, nudim) <- maxhyp(nuplan ,nudim)

                minrgt(nuplan, nudim) <- minhyp(nuplan, nudim)

            END

            maxrgt(nuplan, nudim) <- maxhyp(nuplan, nudim)

        END

    END

END

# 6. Tree homogenous transformation

root :              root of the tree to be transformed

polyed :            list of the polyhedron vertices

minhyp :            list of the polyhedron lower faces

maxhyp :            list of the polyhedron higher faces

dimens :            space dimension

prec1 :             analysis precision

prec2 :             building precision

polybt :            tree of the transformation polyhedron

trfroo :            root of the transformed tree

depth :             computation depth

level :             level reached in the tree

{polroo, minroo, maxroo} :    vertices and faces of the block linked to a node

{pollft, minlft, maxlft} :    left half-polyhedron

{polrgt, minrgt, maxrgt} :    right half-polyhedron

{polblo, minblo, maxblo} :    vertices and faces coming from a terminal node

## 6.1. Computation of the transformed image of a tree according to a homogenous transformation

FUNCTION kdthom(root, polyed, minhyp, maxhyp, dimens ,prec1, prec2)

BEGIN

    kdthom <- *tree* (<u>white</u>)

    polybt <- tree of the polyhedron (polyed, minhyp, maxhyp)

    analysis and non linear transform of the tree (root, kdthom, polybt, polyed, minhyp, maxhyp, dimens, prec1, prec2)

    *delete tree* (polybt)

END

/* Analysis of a tree in order to further compute its transformed release by a non linear transformation*/

PROCEDURE kdanlt(root, trfroo, polybt, polyed, minhyp, maxhyp, dimens, prec1, prec2)

BEGIN

    /* analysis initialization */

    depth <- dimens * prec1, level <- 0

    polroo <- unitary space vertices (dimens)

    minroo <- unitary space lower hyperplanes (dimens)

    maxroo <- unitary space higher hyperplanes (dimens)

    /* recursive computing unit */

    PROCEDURE kdanlt (root, polybt, polroo, minroo, maxroo,[ trfroo, polyed, minhyp, maxhyp,] level)

    BEGIN

      /*search of black terminal nodes included in the polyhedron/*

      IF ((level <> depth) AND (NOT *white* (polybt)) AND (NOT *white* (root)))

      THEN DO

```
IF ((NOT terminal (polybt)) OR (NOT terminal (root)))

THEN DO

    /*parallel depth-first traversal of the tree to be transformed and the tree of
    the polyhedron*/

    {pollft, polrgt} <- vertex-based polyhedron division (polroo)

    {minlft, maxlft, minrgt, maxrgt} <- face-based polyhedron division (minroo,
    maxroo)

    appel kdanlt (left son (root), left son (polybt), pollft, minlft, maxlft, level + 1)

    appel kdanlt (right son (root), right son (polybt), poldrt, minrgt, maxrgt, level
    + 1)

END

ELSE DO

    /* black block included in the polyhedron */

    decomposition of the block in the resulting space(trfroo, polyed, minhyp,
    maxhyp, polroo, minroo, maxroo, dimens, prec2)

END

END

ELSE DO

    /* analysis resolution reached */

    IF ((NOT white (polybt)) AND (NOT white (root)))

    THEN decomposition of the block in the resulting space (trfroo, polyed, minhyp,
    maxhyp, polroo, minroo, maxroo, dimens, prec2)

END

/* tree traversal return path */

delete polyhedron {polroo, minroo, maxroo}

    END

END
```

/* Decomposition of a terminal block in the image space of a geometric transformation */

PROCEDURE kdgtbl(root, polyed, minhyp, maxhyp, polblo, minblo, maxblo, dimens, precis)

BEGIN

    /* decomposition initialization */

    depth <- dimens * precis, level <-0

    /* recursive computing unit */

    PROCEDURE kdtgbl(root, polyed, minhyp, maxhyp,[polblo, minblo, maxblo,] level)

    BEGIN

        intersection evaluation of the terminal block with the transform polyhedron (polblo, minblo, maxblo, polyed, minhyp, maxhyp)

        IF ((level <> depth) AND (NOT *black* (root))AND ((intersection) AND (polyhedron $\not\subset$ block))) THEN DO

            /* the transform polyhedron is not included in the terminal block, it will then divided into two halves */

            {pollft, polrgt} <- vertex-based polyhedron division (polyed)

            {minlft, maxlft, minrgt, maxrgt} <- face-based polyhedron division (minhyp, maxhyp)

            IF (terminal(root)) THEN *fission* (root)

            CALL kdgtbl (*left son* (root), pollft, minlft, maxlft, level + 1)

            CALL kdgtbl (*right son* (root), polrgt, minrgt, maxrgt, level + 1)

        END

        ELSE DO

            /* the polyhedron included in the block is a black node of the transformed tree */

            IF ((NOT *black* (root))AND (intersection)) THEN *blacken* (root)

        END

        /* tree traversal return path */

        IF (NOT *terminal* (root)) THEN *merge* (root)

        <u>IF</u> (level<>0) THEN delete polyhedron(polyed,minhyp, maxhyp)

    END

END

## 6.2. Computation of the transformed image of a tree according to a homogenous transformation (fast transform)

FUNCTION kdfthm(root, polyed, minhyp, maxhyp, dimens, prec1, prec2)

BEGIN

    polybt <- tree of the polyhedron (polyed, minhyp, maxhyp)

    kdfthm <- *tree* (<u>white</u>)

    *value*(kdfthm) <- {polyed, minhyp, maxhyp}

    analysis and non linear transform of the tree, fast transform (racine, kdfthm, polybt, dimens, prec1, prec2)

    *devaluate tree* (kdfthm)

    *delete tree* (polybt)

<u>fin</u>

/* Analysis of a tree in order to further compute its transformed release by a non linear transformation (fast transform)*/

<u>PROCEDURE</u> kdfanl(root, trfroo, polybt, dimens, prec1, prec2)

    This procedure is identical to kdanlt, apart that:

- the transformation polyhedron {polyed ,minhyp, maxhyp} do not more appear explicitly among the call parameters, because it is stored in the root of the transformed tree;

- the block decomposition in the image space is replaced by its fast version.

/* Decomposition of a terminal block in the image space of a geometric transformation (fast version)*/

PROCEDURE kdfgtb(root, polblo, minblo, maxblo, dimens, precis)

BEGIN

    /* decomposition initialization */

    depth <- dimens * precis, level <-0

    /* recursive computing unit */

    PROCEDURE kdtgbr( racine ,[polblo , minblo , maxblo] level)

    BEGIN

        {polyed, minhyp, maxhyp} <- *value*(racine)

        intersection evaluation of the terminal block with the transform polyhedron (polblo, minblo, maxblo, polyed, minhyp, maxhyp)

        IF ((level <> depth) AND (NOT *black* (root)) AND ((intersection) AND (pyhedron $\not\subset$ block))) THEN DO

            /* the transform polyhedron is not included in the terminal block */

            IF (*terminal* (root)) THEN DO

                *merge* (root)

                /*division of the transform polyhedron into two halves, and store them into the root children*/

                {pollft, polrgt} <- vertex-based polyhedron division (polyed)

                {minlft, maxlft, minrgt, maxrgt} <- face-based polyhedron division (minhyp, maxhyp)

                *value*(*left son*(root)) <- { pollft, minlft, maxlft}

                *value*(*right son*(root)) <- { polrgt, minrgt, maxrgt}

            END

            ELSE update the division order of a polyhedron

            CALL kdfgtb (*left son* (root), level + 1)

            CALL kdfgtb (*right son* (root), level + 1)

    END

        ELSE DO

                /* the polyhedron included in the block is a black node of the transformed tree */

                IF ((NOT *black* (root)) AND (intersection)) THEN *blacken* (root)

        END

        /*tree traversal return path*/

        IF (NOT *terminal* (root)) THEN merge and children devaluation (root)

    END

END

# 7. Complements to geometric transforms

| | |
|---|---|
| root : | root of the tree to be transformed |
| symvec : | symmetry axes vector |
| dimens : | dimension of the modeling space |
| precis : | computation precision |
| symaxi : | symmetry axis under analysis |
| depth : | computation depth |
| level : | level reached in the tree |
| dimelm : | dimension of elimination |
| root1 : | root of a minimal sub-tree |
| root2 : | root of a non minimal sub-tree |
| visroo : | visible node in the tree |
| dimvis : | dimension associated to the vision axis |
| lftson : | left son of the root |
| rgtson : | right son of the root |

## 7.1. Computation of the symmetrical tree of a tree

```
FUNCTION kdsymt(root, symvec, dimens, precis)
BEGIN
    /* initialization of the computation parameters */
    depth <- dimens * precis, level <- 0
    symaxi <- link (symvec)
    /* recursive computation unit */
    FUNCTION kdsymt(root, symaxi, level)
    BEGIN
        IF ((NOT terminal (root)) AND (level <> depth)) THEN DO
            /* depth-first tree traversal */
            lftson <- kdsymt(left son (root), link (symaxi), level + 1)
            rgtson <- kdsymt(right son (root), link (symaxi), level + 1)
            /*determining and applying the symmetry*/
            IF (value (symaxi))
            THEN kdsymt <- sub-trees union (rgtson, lftson)
            ELSE kdsymt <- sub-trees union (lftson, rgtson)
        END
        ELSE DO
            /* assertion of the node reached in the tree */
            IF (white (root))
            THEN kdsymt <- tree (white)
            ELSE kdsymt <-   (black)
        END
        merge (kdsymt)
    END
END
```

## 7.2. Removal of hidden parts in a tree along a given dimension

PROCEDURE kdrhpd(root, dimelm, dimens, precis)

BEGIN

    /* initialization of the elimination */

    level <- 0

    /* recursive computation unit */

    PROCEDURE kdrhpd (root,[dimelm,] level)

    BEGIN

        /* looking for nodes couples where the elimination can start */

        IF (NOT *white* (root)) THEN DO

            IF (*terminal* (root)) THEN *fission* (root)

            IF ((level + 1) <> dimelm) THEN DO

                /* depth-first tree traversal looking for initial couples */

                CALL kdrhpd (*left son* (root), level + 1)

                CALL kdrhpd (*right son* (root), level + 1)

            END

            ELSE DO

                /* nodes elimination according to the requested dimension */

                non minimal nodes removal according to the requested dimension (*left son* (root), *right son* (root), dimelm, dimens, precis)

            END

        END

        *merge*(root)

    END

END

## 7.3. Tree traversal with node removal according to the elimination direction

PROCEDURE kdnred(root1, root2, dimelm, dimens, precis)

BEGIN

    /* traversal initialization */

    depth <- precis * dimens, level <- dimelm

    /* recursive computing unit */

    PROCEDURE kdnred (root1, root2, [dimelm, dimens,] level)

    BEGIN

      IF ((level <> depth) AND ((NOT *white* (root1)) OR (NOT *white* (root2)))) THEN DO

        IF (*terminal* (root1)) THEN *fission* (root1)

        IF (*terminal* (root2)) THEN *fission* (root2)

        IF ((MOD(level, dimens) + 1) <> dimelm) THEN DO

          /* direction orthogonal to the elimination direction */

          / *parallel depth-first traversal of the two sub- trees */

          CALL kdnred (*left son* (root1), *left son* (root2), level + 1)

          CALL kdnred (*right son* (root1), *right son* (root2), level + 1)

        END

        ELSE DO

          /* direction parallel to the elimination direction */

          /* depth-first sub-trees traversal and removal of non minimal nodes */

          visroo <- *left son* (root1)

          IF (NOT *white* (visroo)) THEN DO

            IF (NOT *white* (*right son* (root1)))

            THEN CALL kdnred (visroo, *right son* (root1), level + 1)

            IF (NOT *white* (*left son* (root2)))

            THEN CALL kdnred (visroo, *left son* (root2), level + 1)

            IF (NOT *white* (*right son* (root2)))

            THEN CALL kdnred (visroo, *right son* (root2), level + 1)

        END

```
                    visroo <- right son (root1)
                IF (NOT white (visroo)) THEN DO
                    IF (NOT white (left son (root2)))
                    THEN CALL kdnred (visroo, left son (root2), level + 1)
                    IF (NOT white (right son (root2)))
                    THEN CALL kdnred (visroo, right son (root2), level + 1)
                END
                visroo <- left son (root2)
                IF (NOT white (visroo)) THEN DO
                    IF (NOT white (right son (root2)))
                    THEN CALL kdnred (visroo, right son (root2), level + 1)
                END
            END
        END
        ELSE DO
            /* removal of the hidden node */
            IF (NOT white (root1)) THEN whiten (root2)
        END
        /* tree traversal return path */
        merge(root1)
        merge (root2)
    END
END
```

## 7.4. *Accumulation of planes orthogonal with the viewing axis in order to perform a projection*

```
FUNCTION kdplvi(root, dimvis, dimens ,precis)
BEGIN
    /* projection initialization */
    depth <- dimens * precis, level <- 0
    /* recursive computing unit */
    FUNCTION kdplvi(root,[dimvis, dimens,] level)
    BEGIN
        IF ((level <> depth) AND (NOT terminal (root))) THEN DO
            /* depth-first traversal of the initial tree */
            lftson <- kdplvi(left son (root), level + 1)
            rgtson <- kdplvi(right son (root), level + 1)
            /* projection computation */
        IF ((MOD(level, dimens) + 1) = dimvis) THEN DO
            /* projection according to the requested dimension*/
            IF ((white (lftson) OR (white (rgtson))) THEN DO
                IF ((white (lftson))) THEN DO
                    kdplvi <- rgtson
                    delete (lftson)
                END
                ELSE DO
                    kdplvi <- lftson
                    delete tree(rgtson)
                END
            END
```

```
            ELSE DO
                /* union of sub-trees */
                kdplvi <- kdunio(lftson, rgtson, dimens, precis)
                delete (lftson)
                delete (rgtson)
            END
        END
        ELSE sub-trees union (lftson, rgtson)
        /* tree traversal return path */
        merge(kdplvi)
    END
    ELSE DO
         /*assertion of the node reached in the initial tree*/
         IF (NOT white (root))
          THEN kdplvi <- tree (black)
          ELSE kdplvi <- tree (white)
    END
  END
END
```

# 8. Searching adjacencies

root :　　　　root of the tree to be processed

root1, root2 :　　　　symmetrical nodes in the tree

symmat :　　symmetry matrix

dimens :　　space dimension

precis :　　 computation precision

depth :　　computation depth

level :　　level reached in the tree

symma2 :　　 copy of the symmetry matrix

symvec :　　 symmetry vector

symaxi :　　symmetry axis

remain :　　remainder of an Euclidean division

## 8.1. Searching adjacencies among the space objects

PROCEDURE kdsaso(root, symmat, dimens, precis)

BEGIN

    /* search initialization */

    depth <- dimens * precis, level <- 0

    symma2 <- *copy* (symmat)

    /* recursive computing unit */

    PROCEDURE kdsaso (root, symmat, level)

    BEGIN

        IF ((NOT *terminal* (root) AND (level <> depth)) THEN DO

            /* searching adjacencies coming from the reached non terminal block */

            succ <- *link* (symma2)

            WHILE (NOT *nil* (succ)) DO

                /* searching adjacencies according to each vector of the symmetry matrix */

                symvec <- *value* (succ)

                symaxi <- *head extraction* (symvec)

                *queue insertion* (symvec, symaxi)

                IF (symaxi) THEN initialization of the symmetries according to the vector (*left son* (root), *right son* (root), *link* (symvec), dimens, level + 1, depth)

                succ <- *link* (succ)

            END

        /* depth-first tree traversal */

        CALL kdsaso (*left son* (root), *copy* (symma2), level + 1)

        CALL kdsaso (*right son* (root), *copy* (symma2), level + 1)

    END

    *valuate tree*(root)

    delete matrix(symma2)

    END

END

## 8.2. Initialization of the symmetries search according to a given symmetry vector

PROCEDURE kdissv(root1, root2, symvec, dimens, level, depth)

BEGIN

    /* initialization of the search parameters */

    remain <- MOD(level - 1, dimens)

    /* recursive computing unit */

    PROCEDURE kdissv (root1, root2, symvec, [remain,] level)

    BEGIN

        IF (((NOT *terminal* (root1)) OR (NOT *terminal* (root2))) AND ((level <> depth) AND (MOD (level, dimens)<>remain))) THEN DO

            /* depth-first block traversal for building the initial symmetries */

            symaxi <- *value* (symvec)

            IF (symaxi) THEN DO

                /* down move according a symmetry axis: crossing of the nodes */

                CALL kdissv (*right son* (root1), *left son* (root2), *link* (symvec), level + 1)

                CALL kdissv (*left son* (root1), *right son* (root2), *link* (symvec), level + 1)

            END

            ELSE DO

                /* down move orthogonally to a symmetry axis */

                CALL kdissv (*left son* (root1), *left son* (root2), *link* (symvec), level + 1)

                CALL kdissv (*right son* (root1), *right son* (root2), *link* (symvec), level + 1)

            END

        END

        ELSE DO

            /* actual adjacency search */

            searching adjacencies according to a given symmetry vector (root1, root2, symvec, level, depth)

        END

    END

END

## 8.3. Searching symmetries according to a given symmetry vector

PROCEDURE kdsssv(root1, root2, symvec, level, depth)

BEGIN

    /* recursive computing block */

    PROCEDURE kdsssv (root1, root2, symvec, level)

    BEGIN

        IF (((NOT *terminal* (root1)) OR (NOT *terminal* (root2))) AND (level <> depth)) THEN DO

            /* depth-first tree traversal looking for symmetries */

            symaxi <- *value* (symvec)

            IF (symaxi) THEN DO

                /* down move parallel to a symmetry axis*/

                CALL kdsssv (*right son* (root1), *left son* (root2), *link* (symvec), level + 1)

            END

            ELSE DO

                /* down move orthogonally to a symmetry axis*/

                CALL kdsssv (left son (root1), left son (root2), *link* (symvec), level + 1)

                CALL kdsssv (*right son* (root1), *right son* (root2), *link* (symvec), level + 1)

            END

        END

        ELSE DO

            /* precision reached: adjacencies generation for the nodes belonging to space objects*/

            IF ((NOT *white* (root1)) AND (NOT *white* (root2))) THEN DO

                IF (*nil* (*value* (root1))) THEN *value* (root1) <- create an adjacency list

                store an adjacency in a node (*value* (racin1), racin2)

                IF (*nil* (*value* (root2))) THEN value (root2) <- create an adjacency list

                store an adjacency in a node (*value* (racin2), racin1)

            END

        END

    END

END

# 9. Tree labeling and extraction of segment trees

| | |
|---|---|
| root : | root of the labeled tree |
| adclls : | list of the adjacency classes of the tree |
| bucket : | bucket where are stored the adjacent nodes of a component |
| incpls | list of the inside points of a component |
| adjque : | queue of nodes adjacent to a current node |
| cncpnu : | connected component number |
| cncomp : | connected component |
| cptroo : | root of a component tree (segment tree) |
| lftson : | left son of a node |
| rgtson : | right son of a node |

## 9.1. Connected component labeling in a tree

```
FUNCTION kdcclb(root)
BEGIN
    kdcclb <-- create list of connected components
    search of adjacency classes (root, kdcclb)
    tree labeling (kdcclb)
END
```

## 9.2. Search of adjacency classes in atree

```
PROCEDURE kdadcl(root, adclls)
BEGIN
    /* search initialization */
    bucket <-- create a queue
    /* recursive computing unit */

    PROCEDURE kdadcl (root [,adclls])
    BEGIN
        IF ((NOT terminal (root)) AND (nil (value (root)))) THEN DO
            /* depth-first tree traversal looking for adjacency lists */
            CALL kdadcl (left son (root))
            CALL kdadcl (right son (root))
        END
        ELSE DO
            /* adjacency list found */
            IF (NOT nil (value (root))) THEN DO
                /* the node has still not been examined: it is a new connected component */
                incpls <-- create a queue
                insert at queue end (cladjc, incpls)
                insert at queue end (bucket, root)
                connected component analysis (incpls,bucket)
            END
        END
        devaluate tree (root)
    END
    delete queue (bucket)
END
```

## 9.3. Connected component analysis

PROCEDURE kdccan(incpls, bucket)

BEGIN

    /* process adjacent nodes of the bucket until running out */

    WHILE (NOT *empty* (bucket)) DO

        /* move a node from the bucket into the connected component */

        root <-- *extract from queue head* (bucket)

        IF (NOT *nil* (*value* (root))) THEN *insert at queue end* (incpls, root)

        /* scrutation des voisins du nœud transféré */

        filadj <-- *value* (root)

        WHILE (NOT *empty* (adjque)) DO

            /* its neighbors are poured in the bucket */

            *insert at queue end* (bucket, *extract from queue head* (adjque))

        END

        *delete queue*(adjque)

    END

END

## 9.4. Labeling a tree

PROCEDURE kdtlab(adclls)

BEGIN

    /* scrutinizing connected components registered in the adjacency class list */

    cncpnu <-- 0

    cncomp <-- *link* (adclls)

    WHILE (NOT *nil* (cncomp)) DO

        cncpnu <-- cncpnu + 1

        incpls <-- *value* (cncomp)

        /* labeling of the inside of the connected component */

        WHILE (NOT *empty* (incpls)) DO

            root <-- *extract from queue head* (incpls)

            *value* (root) <-- cncpnu

        END

        *delete queue* (incpls)

        erase its references in the structure (cncomp)

        cncomp <-- *link* (cncomp)

    END

END

## 9.5. Building the segment trees from the connected components of a labeled tree

PROCEDURE kdbsgt(cladjc,racine)

BEGIN

    /* scrutinizing the connected components registered in the adjacency class list */

    cncpnu <-- 0

    cncomp <-- *link* (adclls)

    WHILE (NOT *nil* (cncomp)) DO

        /* building the tree of the current connected component */

        cncpnu <-- cncpnu + 1

        cptroo <-- extraction of a connected component from a tree (root, cncpnu)

        /* hanging of the segment tree in the adjacency class list */

        *value* (cncomp) <-- cptroo

    END

END

## 9.6. Extraction of a connected component from a tree

FUNCTION kdexcc(root, cncomp)

BEGIN

    /* recursive computing unit */

    FUNCTION kdexcc(root [,cncomp])

    BEGIN

        IF ((NOT *terminal* (root)) AND (*nil* (*value* (root))))

        THEN DO

            /* depth-first tree traversal searching labeled nodes */

            lftson <-- kdexcc (*left son* (root))

            rgtson <-- kdexcc (*right son* (root))

            kdexcc <-- *sub-trees union* (lftson, rgtson)

        END

        ELSE DO

            /*create a black node when the node label is corresponding to the searched component*/

            IF (*value* (root) = cncomp)

            THEN kdexcc <-- *tree* (<u>black</u>)

            ELSE kdexcc <-- *tree* (<u>white</u>)

        END

        /* tree traversal return path */

        IF ((NOT *terminal* (kdexcc)) THEN *merge* (kdexcc)

    END

END

# 10. Computation of the generalized moment list of a tree

root :               root of the tree

dimens :             dimension of the modeling space

precis :             computation precision

depth :              computation depth

level :              level reached in the tree

minroo :             minimum coordinates list of the root

maxroo :             maximum coordinates list of the root

{ minson, maxson} :  min. et max. coordinates list of a son

nudim :              coordinate number

{nudim1, nudim2, nudim3} :   coordinate numbers

{x1 , x2, x3} :      simple, squared, cubic value of a coordinate

x :                  coordinate

ncoord :             number of the block dividing coordinate

moment :             moment list

index :              computation or table index

momroo:              moment list of the root

{momlft, momrgt} :   moment lists of left and right sons

## 10.1. Computation of the moment list of a tree

FUNCTION kdmomt (root, dimens, precis)

BEGIN

/* moments computation initialization */

{minroo, maxroo} <-- vector {(0., 0., ..., 0.),(1.,1., ..., 1.)}

momroo <-- moment list of the unitary space (dimens)

value(root) <-- momroo

depth <-- precis * dimens,  level <-- 0

/* recursive computing unit */

PROCEDURE kdmomt (root, minroo, maxroo, level)

BEGIN

IF ((NOT *terminal* (root)) AND (level <> depth)) THEN DO

/* computation of the left node moments */

minson <-- *copy* (minroo),  maxson <-- *copy* (maxroo)

nudim <-- MOD(level, dimens) + 1

momroo <-- *value* (root)

*value* (*left son* (root)) <-- child moment list computation (momroo, nudim, minroo(nudim), dimens)

maxson(nudim) <-- (minroo(nudim) + maxroo(nudim))/2.

CALL kdmomt (*left son* (root), minson, maxson, level + 1)

/* computation of the right node moments */

minson <-- *copy* (minroo), maxson <-- *copy* (maxroo)

nudim <-- MOD(level, dimens) + 1

momroo <-- *value* (root)

*value* (*right son* (root)) <-- child moment list computation (momroo, nudim, maxroo(nudim), dimens)

minson(nudim) <-- (minroo(nudim) + maxroo(nudim))/2.

CALL kdmomt (*right son* (root), minson, maxson, level + 1)

/* evaluation of the tree moments on return path */

cumulative children moments (root)

END
    END

    /* moment list retrieval */
    kdmomt <-- *value* (root)
    *value*(root) <-- nil
END

## 10.2. Initialization of the moment list of the unitary space

```
FUNCTION kdimus(dimens)
BEGIN
    /* computation of the order 0 moment */
    kdimus (0, 0, 0) <-- 1.
    /* computation of the order 1 moments */
    FOR nudim1=1 TO dimens DO
        kdimus (nudim1, 0, 0) <-- 1./2.
    END
    /* computation of the order 2 moments */
    FOR nudim1=1 TO dimens DO
        FOR nudim2 = nudim1 TO dimens DO
            IF (nudim1 = nudim2)
            THEN kdimus (nudim1, nudim2, 0) <-- 1./3.
            ELSE kdimus (nudim1, nudim2, 0) <-- 1./4.
        END
    END
    /* computation of the order 3 moments */
    FOR nudim1 = 1 TO dimens DO
        FOR nudim2 = nudim1 TO dimens DO
            FOR nudim3 = nudim2 TO dimens DO
                IF ((nudim1 = nudim2) AND (nudim2 = nudim3))
                THEN kdimus (nudim1, nudim2, nudim3) <-- 1./4.
                ELSE DO
                    IF ((nudim1 = nudim2) OR (nudim2 = nudim3))
                    THEN kdimus (nudim1, nudim2, nudim3) <-- 1./6.
                    ELSE kdimus (nudim1, nudim2, nudim3) <-- 1./8.
                END
            END
        END
    END
```

END

## 10.3. Computation of a child moment list

```
FUNCTION kdchml(moment, ncoord, x, dimens)
BEGIN
    x1 <-- x, x2 <--x1*x, x3 <--x2*x
    /* computation of the order 0 moment */
    kdclmg(0, 0, 0) <-- moment(0, 0, 0)/2.
    /* computation of the order 1 moment */
    FOR nudim1=1 TO dimens DO
        IF (nudim1 = ncoord)
        THEN kdchml (nudim1, 0, 0) <-- (moment(nudim1, 0 ,0)/4.) + ((moment(0, 0, 0)*x1)/4.)
        ELSE kdchml (nudim1 ,0, 0) <-- moment(nudim1, 0, 0)/2.
    END
    /* computation of the order 2 moment */
    FOR nudim1 = 1 TO dimens DO
        FOR nudim2 = nudim1 TO dimens DO
            IF ((nudim1 <> ncoord) AND (nudim2 <> ncoord))
            THEN kdchml (nudim, nudim2, 0) <-- moment(nudim1, nudim2, 0)/2.
            ELSE DO
                IF ((nudim1 <> ncoord) OR (nudim2 <>n coord)) THEN DO
                    IF (nudim1 <> ncoord) THEN kdchml (nudim1, nudim2, 0)
                        <-- (moment(nudim1, nudim2,0)/4.)+ ((moment(nudim1, 0, 0)*xl)/4.)
                    ELSE kdchml (nudim1 ,nudim2,0)
                        <-- (moment(nudim1, nudim2,0)/4.)+ ((moment(nudim2 ,0,0)*xl)/4.)
                END
                ELSE kdchml (nudim1, nudim2, 0) <-- (moment(nudim1, nudim2, 0)/8.)
                    + ((moment(nudim1, 0, 0)*xl)/4.)+ ((moment(0, 0 , 0)*x2)/8.)
            END
        END
    END
```

```
/* computation of the order 3 moment */
FOR nudim1=1 TO dimens DO
    FOR nudim2=nudim1 TO dimens DO
        FOR nudim3=nudim2 TO dimens DO
            index <---0
            IF (nudim1 = ncoord) THEN index <-- index+4
            IF (nudim2 = ncoord) THEN index <-- index+2
            IF (nudim3 = ncoord) THEN index <-- index+1
            ACCORDING TO (index) DO
            0 : kdchml (nudim1, nudim2, nudim3)
                    <-- moment(nudim1, nudim2, nudim3)/2.
            1, 2, 4 : BEGIN
                    IF (index = 1) THEN kdchml (nudim1, nudim2, nudim3)
                        <-- (moment(nudim1, nudim2, nudim3)/4.)
                        +  ((moment(nudim1, nudim2, 0)*xl)/4.)
                    IF (index = 2) THEN kdchml (nudim1, nudim2, nudim3)
                        <-- (moment(nudim1, nudim2, nudim3)/4.)
                        +  ((moment(nudim1, nudim3, 0)*xl)/4.)
                    IF (index = 4) THEN kdchml (nudim1, nudim2, nudim3)
                        <-- (moment(nudim1, nudim2, nudim3)/4.)
                        +  ((moment(nudim2, nudim3, 0)*xl)/4.)
                END
```

```
                3,5,6 : BEGIN

                        IF (index = 3) THEN kdchml (nudim1, nudim2, nudim3)

                            <-- (moment(nudim1, nudim2, nudim3)/8.)

                            + ((moment(nudim1, nudim2, 0)*x1)/4.)

                            + ((moment(nudim1 0, 0)*x2)/8.)

                        IF (index = 5) THEN kdchml (nudim1, nudim2, nudim3)

                            <-- (moment(nudim1, nudim2, nudim3)/8.)

                            + ((moment(nudim2, nudim3, 0)*x1/4.)

                            + ((moment(nudim2, 0, 0)*x2)/8.)

                        IF (index = 6) THEN kdchml (nudim1, nudim2, nudim3)

                            <-- (moment(nudim1, nudim2, nudim3)/8.)

                            + ((moment(nudim2, nudim3, 0)*x1)/4.)

                            + ((moment(nudim3, 0, 0)*x3)/8.)

                END

                7 : kdchml (nudim1, nudim2, nudim3)

                        <-- (moment(nudim1, nudim2, nudim3)/16.)

                        + ((moment(nudim1, nudim2, 0)*x1)*(3./16.))

                        + ((moment(nudim1 ,0, 0)*x2)*(6./32.))

                        + ((moment(0, 0, 0)*x3)/16.)

                END

            END

        END

    END

END
```

## 10.4. Cumulative node children moments

PROCEDURE kdcucm(root)

BEGIN

    /* retrieval of father and children moment lists */

    momroo <-- *value* (root)

    momlft <-- *value* (*left son* (root))

    momrgt <-- *value* (*right son* (root))

    /* scrutinizing of the three lists and cumulative children moments into father's ones */

    FOR index = 0 TO ($C^1_{dimens} + C^2_{dimens} + C^3_{dimens}$) DO

        IF (*white* (*left son* (root))) THEN momlft(index) <-- 0.

        IF (*white* (*right son* (root))) THEN monrgt(index) <-- 0.

        momroo(index) <-- momlft(index) + momrgt(index)

    END

    /* devaluation of the children nodes */

    *delete list* (momlft)

    *delete list* (momrgt)

    *value* (*left son* (root)) <-- <u>nil</u>

    *value* (*right son* (root)) <-- <u>nil</u>

END

# 11. Centering and normalizing a generalized moments list

| | |
|---|---|
| moment : | moment list to be processed |
| momnor : | normalized moment list |
| matrot : | generalized rotation matrix in homogenous coordinates |
| dimens : | dimension of the modeling space |
| nudim : | coordinate number |
| nudim1 : | first coordinate number |
| nudim2 : | second coordinate number |
| nudim3 : | third coordinate number |
| vectr : | vector of the gravity center |
| hypvol : | tree hypervolume |
| matsym : | inertia matrix extracted from the moment list |
| matvpr : | matrix of Eigen vectors of the tree Eigen reference frame |

## 11.1. Generation of the centered list from a moment list

```
FUNCTION kdctrm(moment, dimens)
BEGIN
    /* evaluation of the tree hypervolume */
    hypvol <-- moment(0 ,0, 0)
    /* storing the hypervolume */
    kdctrm(0, 0 ,0) <-- hypvol
    /* evaluation of the gravity center */
    FOR nudim1=1 TO dimens DO
        kdctrm(nudim1, 0, 0) <-- moment(nudim1, 0, 0)/hypvol
        vectr(nudim1) <-- kdctrm(nudim1, 0, 0)
    END
    /* centering the order 2 moments */
    FOR nudim1=1 TO dimens DO
        FOR nudim2=nudim1 TO dimens DO
            kdctrm(nudim1, nudim2, 0) <-- moment(nudim1, nudim2, 0)
            - (vectr(nudim2)*moment(nudim1, 0, 0))
            -  (vectr(nudim1)*moment(nudim2, 0, 0))
            +  (vectr(nudim1)*vectr(nudim2)*hypvol)
            kdctrm(nudim1, nudim2, 0)
            <-- kdctrm(nudim1, nudim2, 0)/hypvol
        END
    END
```

```
/* centering the order 3 moments */
FOR nudim1=1 TO dimens DO
    FOR nudim2=nudim1 TO dimens DO
        FOR nudim3=nudim2 TO dimens DO
            kdctrm(nudim1, nudim2, nudim3)
                <-- moment(nudim1, nudim2, nudim3)
                 - (vectr(nudim3)*moment(nudim1, nudim2, 0))
                 - (vectr(nudim2)*moment(nudim1, nudim3, 0))
                 - (vectr(nudim1)*moment(nudim2, nudim3, 0))
                 + (vectr(nudim2)*vectr(nudim3)*moment(nudim1, 0, 0))
                 + (vectr(nudim1)*vectr(nudim3)*moment(nudim2, 0, 0))
                 + (vectr(nudim1)*vectr(nudim2)*moment(nudim3, 0, 0))
                 - (vectr(nudim1)*vectr(nudim2)*vectr(nudim3)*hypvol)
            kdctrm(nudim1, nudim2, nudim3)
                <-- kdctrm(nudim1, nudim2, nudim3)/hypvol
        END
    END
END
END
```

## 11.2. Generation of the moment normalized list and the rotation matrix

PROCEDURE kdnrmr(moment, momnor, matrot, dimens)

BEGIN

    /* copying out volume and gravity center coordinates*/

    momnor(0, 0, 0) <-- moment(0, 0, 0)

    FOR nudim1=1 TO dimens DO

        momnor(nudim1, 0, 0) <-- moment(nudim1, 0, 0)

    END

    /* building and diagonalization of the inertia matrix */

    FOR nudim1=1 TO dimens DO

        FOR nudim2=nudim1 TO dimens DO

            matsym(nudim1, nudim2) <-- moment(nudim1, nudim2, 0)

        END

    END

    compute Eigen values and vectors (matsym, matvpr, dimens)

    /* inertia axes computation */

    FOR nudim=1 TO dimens DO

        momnor(nudim, nudim, 0) <-- matsym(nudim, nudim)

    END

```
/* asymmetries computation */
FOR nudim=1 TO dimens DO
    momnor(nudim, nudim, nudim) <-- 0.
    FOR nudim1=1 TO dimens DO
        FOR nudim2=1 TO dimens DO
            FOR nudim3=1 TO dimens DO
                momnor(nudim, nudim, nudim)
                    <-- momnor(nudim, nudim, nudim)
                    + (moment(nudim1, nudim2, nudim3)
                    *matvpr(nudim ,nudim1)
                    *matvpr(nudim, nudim2)
                    *matvpr(nudim, nudim3))
            END
        END
    END
    /* reduce the asymmetry dimensionality */
    momnor(nudim ,nudim, nudim)
        <-- CUBIC ROOT (momnor(nudim ,nudim, nudim))
END
matrot <--matrix conversion in homogenous coordinates (matvpr)
END
```

## 11.3. Generation of the moment normalized list

FUNCTION kdnrmg(moment, dimens)

This function is analogous to kdnrmr, except that:

- the generalized rotation matrix is not retrieved;
- the uncertainty about the directions of Eigen axes is removed giving back positive asymmetries.

## 12.     Generation of an Eigen tree

root :                  root of the tree to be transformed

momnor :                tree centered and reduced moment list

matrot :                rotation matrix associated to the Eigen reference frame

dimens :                dimension of the modeled space

prec1 :                 analysis precision

prec2 :                 computation precision

nudim :                 coordinate number

vect :                  work vector

{mdtr, mitr }           direct and inverse translation matrices

{mdan, mian} :          direct and inverse homothety matrices

{mdcr, micr} :          direct and inverse rotation centering matrices

mirt :                  inverse rotation matrices

{mart ,mtra} :          direct and inverse transformation matrices

 {polesp, minesp, maxesp} :     unitary space polyhedron

{ polyed, minhyp, maxhyp} :     transformation polyhedron

homotr :                homogenous transformed tree

tabsym :                table of symmetries (signature of the Eigen reference frame)

vecsym :                symmetry vector

## 12.1. Generation of the Eigen tree of a tree

FUNCTION kdeigt(root, momnor , matrot, dimens, prec1, prec2)

BEGIN

    /* building of the direct translation matrix */

    FOR nudim=1 TO dimens DO

        vect(nudim) <-- 0.5-momnor(nudim, 0, 0)

    END

    mdtr <-- matrix of translation(vect)

    /* computation of the direct homothety matrix */

    FOR nudim=1 TO dimens DO

        vect(nudim) <--1./(momnor(1, 1, 0)*6.)

    END

    mdan <-- matrix of anamorphose (vect)

    /* computation of the rotation centering matrix */

    FOR nudim=1 TO dimens DO

        vect(nudim) <-- 0.5

    END

    mdcr <-- matrix of translation (vect)

    micr <-- opposite (mdcr)

    /* computation of the direct transformation matrix */

    mart <-- concaténation des matrices (mdan, mdcr, matrot, micr, mdtr)

    /* computation of the inverse matrices */

    mian <-- inverse(mdan)

    mirt <-- transposed (matrot)

    mitr <-- opposite(mdcr)

    /* computation of the inverse transformation matrix */

    mtra <-- concatenation of the matrices(mitr, mdcr, mirt, mian, micr)

    /* building of the transformation polyhedron from its inverse matrix */

    polesp <- vertex-based polyhedron of the unitary space (dimens)

    polyed <-- polyhedron transformation (polesp, mtra)

```
/* building of the lower and higher hyperplanes of the transformation polyhedron from its
direct matrix */

{minesp,maxesp} <-- face-based polyhedron of the unitary space (dimens)

{minhyp,maxhyp} <-- hyperplanes transformation ({minesp, maxesp}, mart)

/* generation of the Eigen tree */

homotr  <-- tree homogenous transformation (racine, polyed, minhyp ,maxhyp, dimens,
prec1, prec2)

/* evaluation of the signature of the Eigen reference frame */

FOR nudim=1 TO dimens DO
        IF (momnor(nudim, nudim ,nudim)<0.)
        THEN tabsym(nudim) <-- true
        ELSE tabsym(nudim) <-- false
END

/* reversal of the Eigen tree */

vecsym <-- Boolean vector (tabsym)

kdeigt <-- symmetrical tree of the tree(homotr , vecsym, dimens, prec2)

END
```

## 12.2. Generation of the Eigen tree of a tree (fast version)

FUNCTION kdfeit(root, momnor, matrot, dimens, precl, prec2)

This function is analogous to kdeigt, except that:

- the tree homogenous transform is replaced by its fast version.